\documentclass[lettersize,journal]{IEEEtran}
\usepackage{amsmath,amsfonts}
\usepackage{algorithmic}
\usepackage{algorithm}
\usepackage{array}
\usepackage[caption=false,font=normalsize,labelfont=sf,textfont=sf]{subfig}
\usepackage{textcomp}
\usepackage{stfloats}
\usepackage{url}
\usepackage{verbatim}
\usepackage{graphicx}
\usepackage{cite}
\usepackage{capt-of}
\usepackage{cite}
\usepackage{bm}
\usepackage{makecell}
\usepackage{xspace}
\usepackage{amsmath} 
\usepackage{amssymb}  
\usepackage[table,dvipsnames,xcdraw]{xcolor}
\usepackage{amsfonts}
\usepackage{enumitem, multicol} 

\usepackage{graphics} 
\usepackage{epsfig} 
\usepackage{mathptmx} 
\usepackage{times} 

\usepackage{mathrsfs}
\usepackage{indentfirst}
\usepackage{multirow}
\usepackage{adjustbox}

\usepackage{color}      
\usepackage{pifont}     
\usepackage{algorithmic}
\usepackage{graphicx}
\usepackage{textcomp}
\usepackage{bbding}

\usepackage{colortbl} 
\usepackage{diagbox} 
\usepackage{siunitx}
\usepackage{float}

\usepackage[colorlinks=true,linkcolor=blue,citecolor=green,urlcolor=blue,]{hyperref}
\usepackage{orcidlink}

\usepackage{algorithm}
\usepackage{algorithmic}

\usepackage{etoolbox}
\usepackage[compact]{titlesec}

\titlespacing*{\section}{0pt}{0.55\baselineskip plus 0.1\baselineskip minus 0.1\baselineskip}{0.25\baselineskip}
\titlespacing*{\subsection}{0pt}{0.45\baselineskip plus 0.1\baselineskip minus 0.1\baselineskip}{0.15\baselineskip}
\titlespacing*{\subsubsection}{0pt}{0.35\baselineskip plus 0.1\baselineskip minus 0.1\baselineskip}{0.10\baselineskip}

\makeatletter
\patchcmd{\@makecaption}
  {\@IEEEtabletopskipstrut{\normalfont\footnotesize #1}\\{\normalfont\footnotesize\scshape #2}}
  {\@IEEEtabletopskipstrut{\parbox[t]{\hsize}{\centering\normalfont\footnotesize #1.\nobreakspace\scshape #2}}}
  {}{}
\makeatother

\colorlet{colorFst}{Green!25}       
\colorlet{colorSnd}{SpringGreen!45} 
\colorlet{colorTrd}{Yellow!30}      
\colorlet{colorLow}{darkgray!30}    
\definecolor{R1}{HTML}{E97451}
\definecolor{R2}{HTML}{008080}
\definecolor{R3}{HTML}{0047AB}
\colorlet{cmt}{darkgray!80}    
\colorlet{supp}{darkgray!50}    

\newcommand{\blackx}{{\color{black}\ding{55}}}

\newcommand{\fs}{\cellcolor{colorFst}}   
\newcommand{\nd}{\cellcolor{colorSnd}}      
\newcommand{\rd}{\cellcolor{colorTrd}}      

\begin{document}

\definecolor{nircolor}{RGB}{0,100,150} 
\definecolor{m3dgrcolor}{RGB}{204,112,0}  
\newcommand{\ultrafusion}{Ultra-Fusion\xspace}
\newcommand{\mthreeDGR}{M3DGR\xspace}
\newcommand{\ultrafusionc}{\texorpdfstring{\textcolor{nircolor}{Ultra-Fusion}\xspace}{Ultra-Fusion}}
\newcommand{\mthreeDGRc}{\texorpdfstring{\textcolor{m3dgrcolor}{M3DGR}\xspace}{M3DGR}}

\definecolor{darcolor}{RGB}{0,100,150}    

\title{\LARGE 
\bf{\ultrafusionc}: A Resilient Tightly-Coupled Multi-Sensor Fusion SLAM Framework under Sensor Degradation and Spatiotemporal Perturbation for Intelligent Transportation Systems

}

\IEEEaftertitletext{%
\vspace{-3\baselineskip}
\begin{center}
\large\textbf{Project webpage: \url{https://sjtuyinjie.github.io/ultrafusion-web/}}
\end{center}
\vspace{0.15\baselineskip}
\begin{center}
\includegraphics[width=0.75\linewidth]{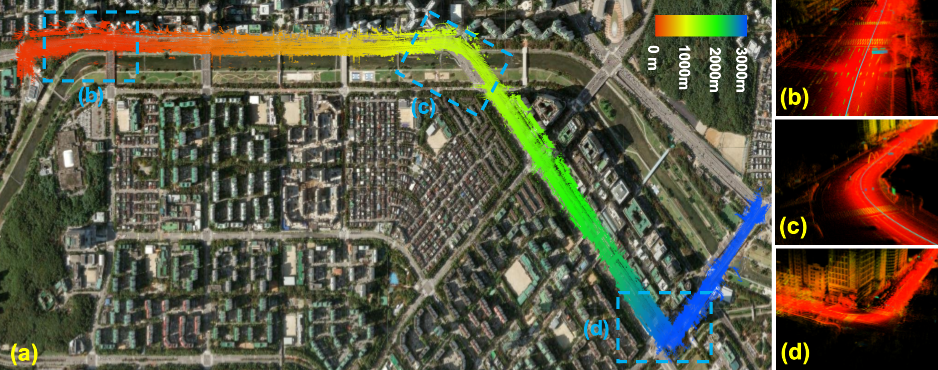}
\end{center}
\vspace{-0.5mm}
{\small
\refstepcounter{figure}\label{fig:kaist_overview}
\noindent{Fig.~\thefigure.}~\ultrafusion supports road-level localization and reconstruction for urban autonomous driving.
(a)~A 3~km KAIST urban-driving route, color-coded by traveled distance.
(b)--(d)~Local reconstructions along the route, preserving road boundaries, roadside structures, and the estimated vehicle trajectory (blue).
\par
}
\vspace{0.1\baselineskip}
}

\author{Yihong Tian~$^{1}$, Junjie Zhang~$^{2}$, Liuyang Li~$^{3}$, Deteng Zhang~$^{4}$, Yunfei Zuo$^{1}$ and Jie Yin$^{5}$* 
\thanks{$^*$ Corresponding author: Jie Yin ({\small\url{robot_yinjie@outlook.com}}). }
\thanks{$^{1}$Beijing Institute of Technology, $^{2}$Chongqing University,
$^{3}$Sichuan University, $^{4}$Northwestern Polytechnical University, $^{5}$Shanghai Jiao Tong University}%
}



\maketitle

\begin{abstract}

Reliable localization is essential for intelligent transportation systems (ITS), including autonomous vehicles, quadruped last-mile carriers, and infrastructure-inspection unmanned aerial vehicles (UAVs). Although tightly-coupled multi-sensor fusion improves accuracy in favorable conditions, deployed systems remain vulnerable to sensor degradation---poor illumination, LiDAR degeneracy, wheel slippage, and GNSS outage---and to spatiotemporal calibration errors. These failures are common in urban canyons, tunnels, and high-speed corridors, where localization drift can degrade route tracking, tunnel passage continuity, and local map alignment.
This paper presents \ultrafusionc, a tightly-coupled multi-sensor localization framework based on a unified sliding-window estimator. Asynchronous measurements are timestamp-ordered and converted into optional factors within one optimization window, supporting WIO, VIO, LIO, and LVIO with optional wheel and GNSS augmentation. Observability-aware initialization selects the bootstrap mode, factor-wise reliability scheduling gates degraded measurements, and online LiDAR--IMU spatiotemporal calibration refines temporal offsets and rotational extrinsics during operation.
We extend the \mthreeDGRc benchmark with simulation trajectories and evaluate more than 60 open-source SLAM systems on \mthreeDGR, M2DGR-Plus, KAIST, GrandTour, and MARS-LVIG. The results show competitive accuracy across wheeled, legged, and aerial platforms under long-duration and high-speed operation, degradation, and calibration perturbation, improving localization availability for road-level autonomy, campus and warehouse mobility, and low-altitude aerial inspection.
To benefit the industrial and academic community, we will release source code and datasets upon paper acceptance.

\end{abstract}

\begin{IEEEkeywords}
System State Estimation, Sensor Technology, Connected and Autonomous Vehicles, Autonomous Driving, Multi-sensor Fusion, System Design
\end{IEEEkeywords}

\begin{figure*}[t]
    \centering
    \includegraphics[width=1.8\columnwidth]{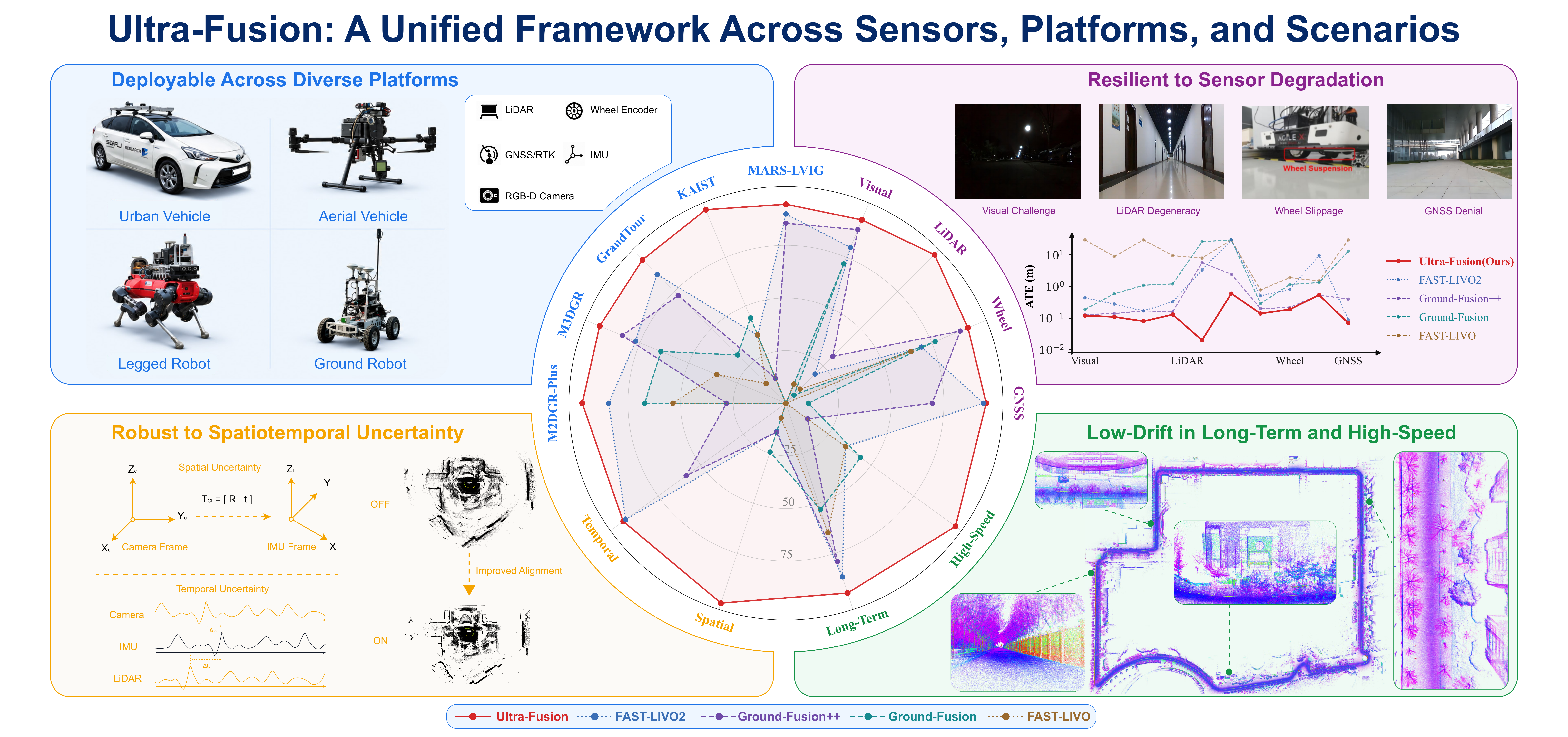}\\
    \caption{Overview of Ultra-Fusion as a unified multi-sensor SLAM framework across sensors, platforms, and scenarios. The framework supports heterogeneous inputs, deploys on ground, aerial, legged, and vehicle platforms, and improves localization robustness under sensor degradation, spatiotemporal uncertainty, and long-term or high-speed operation.}
    \label{fig:teaser}
    \vspace{-4mm}
\end{figure*}

\section{Introduction}

\IEEEPARstart{R}{eliable} localization is central to intelligent transportation systems, including autonomous vehicles, advanced driver-assistance systems (ADAS), infrastructure-assisted mobility, quadruped delivery and inspection platforms, and low-altitude UAV operations~\cite{al2024review,yin2023design}. In urban streets, highways, tunnels, campuses, and air corridors, localization must handle changing sensing conditions, platform motions, and environmental uncertainty. SLAM has therefore evolved from single-modality pipelines toward multi-sensor fusion with LiDAR, cameras, GNSS, wheel odometry, and IMUs. Although existing tightly- and loosely-coupled frameworks perform well in structured settings~\cite{lin2021r3live,yin2023sky,qu2024implicit}, robustness across sensor configurations, platforms, and operating conditions remains challenging.

This gap reflects several coupled requirements. First, ITS fleets rarely share the same sensor suite: road vehicles and shuttles may rely on wheel odometry and intermittent GNSS, quadruped platforms encounter slip and body oscillation on uneven terrain, and UAVs face weak structure at altitude, whereas each may combine different visual, LiDAR, and inertial sensors. Fixed-stack systems often require changes to state definitions, initialization, factor activation, calibration variables, or marginalization when the sensor suite changes. Second, sensor reliability varies over time. Poor illumination, dynamic occlusion, motion blur, adverse weather, LiDAR degeneracy, wheel slippage, and GNSS denial can make useful measurements unreliable; fixed-confidence fusion may then bias the optimization or cause tracking loss. Third, spatiotemporal inconsistency is common in multi-sensor platforms, especially in LVIO, where time offsets and extrinsic errors distort cross-modal correspondences and accumulate drift~\cite{khedekar2022mimosa,yin2024ground}.
Based on these observations, we argue that ITS localization should be treated as a unified estimation problem rather than as configuration-specific pipelines. A desirable framework should support multiple sensor configurations, initialize under sufficient observability, adapt degraded factors online, and refine calibration only under reliable excitation. Evaluation should also go beyond a single complete sensor suite, since robustness depends on whether each configuration remains competitive within its method category and transfers across representative platforms and scenarios.

To address these requirements, we propose Ultra-Fusion, a tightly-coupled multi-sensor framework for ITS localization under flexible sensor availability. A \emph{Unified Sliding-Window Estimator} orders heterogeneous observations by timestamp and converts them into compatible factors in one optimization window, supporting WIO, VIO, LIO, and LVIO with optional wheel and/or GNSS augmentation. \emph{Observability-Aware Initialization} selects the start-up regime, \emph{Factor-Wise Reliability Scheduling} gates or down-weights unreliable residuals, and \emph{Online Spatiotemporal Calibration} refines temporal and extrinsic parameters under sufficient excitation and sensor reliability.

\ultrafusion provides a \textbf{unified framework across sensors, platforms, and scenarios}: WIO, VIO, LIO, LVIO and their variants share initialization, reliability scheduling, calibration, and marginalization logic. We evaluate the estimator under different sensor availability, platform dynamics, and operating conditions, including simulation-based perturbations and an expanded \mthreeDGR comparison. The main contributions are:
\begin{itemize}

\item We propose \textbf{\ultrafusionc\footnote{\url{https://github.com/sjtuyinjie/Ultra-Fusion}}}, a tightly-coupled multi-sensor fusion framework for intelligent transportation systems. A unified sliding-window estimator with observability-aware initialization supports WIO, VIO, LIO, and LVIO within one configurable optimization framework that shares state representation, factor admission, calibration, and marginalization.

\item We develop \textbf{Factor-Wise Reliability Scheduling} to address sensor degradation during online estimation. The scheduler applies degeneracy-aware gating and down-weighting directly within the unified optimization problem, improving robustness under poor illumination, LiDAR degeneracy, wheel slippage, and GNSS denial.

\item We design \textbf{Online Spatiotemporal Calibration} to handle calibration uncertainty during operation. The calibration variables are refined only under sufficient excitation and sensor reliability, reducing cross-modal bias caused by temporal offsets and extrinsic perturbations.

\item We extend the \textbf{\mthreeDGRc benchmark\footnote{\url{https://github.com/sjtuyinjie/M3DGR}}} and conduct a comprehensive large-scale evaluation of \textbf{more than 60} representative SLAM systems, including controlled simulation-based perturbation studies. This benchmark study provides a systematic analysis of robustness trends, degradation patterns, and common failure modes.

\item We validate \ultrafusion on heterogeneous ITS-relevant platforms and operating conditions, including autonomous driving, campus and warehouse wheeled robots, quadruped robots, and inspection UAVs, as well as long-term and high-speed trajectories.

\end{itemize}

Together, the framework and benchmark support deployment-oriented multi-sensor localization under sensing degradation and calibration uncertainty across multimodal transportation platforms.

\begin{table*}[ht] 
\small 
\caption{Comparison of representative multi-sensor fusion SLAM systems, emphasizing configurability and robustness.} 
\centering 
\renewcommand{\arraystretch}{1.3} 
\label{slam_comparison} 
\begin{adjustbox}{width=1.8\columnwidth} 
\begin{tabular}{*{7}c} 
\hline 
\makecell{{Method/Year}} & \makecell{Sensor Set$^1$} & \makecell{Configurable\\Modes} & {Tightly-Coupled} & Degradation-aware & \makecell{Online\\Calibration} & Mapping$^2$ \\ 
\hline 
VINS-RGBD\cite{shan2019rgbd}, 2019 & CID & Fixed & \Checkmark & & & SPC/Non-color\\ 
DRE-SLAM\cite{yang2019dre}, 2019 & CDW & Fixed & \Checkmark & & & Mesh/Non-color\\ 
GR-Fusion\cite{wang2021gr}, 2021 & CIWGL & Fixed & \Checkmark & & & SPC/Non-color\\ 
LVI-SAM\cite{10.1109/ICRA48506.2021.9561996}, 2021 & CIL & Fixed &  & & & SPC/Non-color\\ 
VIW-Fusion\cite{Tingda2022VIW}, 2022 & CIW & Fixed & \Checkmark & & Spatial+Temporal  & SPC/Non-color\\ 
R3LIVE\cite{lin2021r3live}, 2022 & CIL & Fixed & \Checkmark & & & \fs \bf{Mesh/Color}\\ 
DAMS-LIO\cite{han2023dams}, 2023 & IWL & Fixed & \Checkmark & L & & {SPC/Non-color}\\ 
M2C-GVIO\cite{hua2023m2c}, 2023 & CIG & Fixed & \Checkmark & & & Sparse/Non-color\\ 
FAST-LIVO2\cite{zheng2024fast}, 2024 & CIL & Fixed & \Checkmark & L& & \nd\sl{DPC/Color}\\
Ground-Fusion\cite{yin2024ground}, 2024 & CIDWG & GNSS-optional & \Checkmark & CDWG& Spatial & \nd \sl{DPC/Color}\\ 

LIGO\cite{he2025ligo}, 2025 & IGL & Fixed & \Checkmark & & & SPC/Non-color\\ 
Super Odometry\cite{zhao2025resilient}, 2025 & CIL & Fixed & \Checkmark & L & & SPC/Non-color\\

{Ground-Fusion++}\cite{zhang2025towards}, 2025 & CIDWGL & GNSS-optional &  & L & & \fs \bf{Mesh/Color}\\ 

\hline 
\textbf{\ultrafusion(Ours)}, 2026 & CIDWGL & \textbf{WIO \& VIO/LIO/LVIO + optional W/G} & \Checkmark & CIDWGL & Spatial+Temporal & \fs \bf{GS/Color}\\ 
\hline 
\multicolumn{7}{l}{\footnotesize{$^1$ C: RGB camera, I: IMU, D: depth camera, W: wheel odometry, G: GNSS, L: LiDAR.$^2$ SPC: Sparse Point Cloud, DPC: Dense Point Cloud, GS: Gaussian Splatting. }} 
\end{tabular} 
\end{adjustbox} 
\vspace{-5mm}
\end{table*}

\section{Related Work}

\subsection{Multi-sensor Fusion SLAM}

Multi-sensor fusion SLAM exploits complementary measurements to balance accuracy, drift, and robustness. Visual and visual--inertial systems are effective in texture-rich scenes~\cite{campos2021orb,von2022dm,qin2018vins}, while LiDAR-centric pipelines provide geometric observability and scale consistency~\cite{shan2020lio,xu2022fast,bai2022faster}. Recent LiDAR--visual--inertial systems combine these cues for pose estimation and mapping~\cite{lin2021r3live,zheng2022fast,zheng2024fast}. In ITS, road vehicles, quadruped carriers, and inspection UAVs often add wheel odometry and GNSS for short-term motion constraints and long-term drift correction~\cite{Tingda2022VIW,cao2022gvins,he2025ligo,hua2023m2c,li2026p}.
A key design choice is how LiDAR geometry enters the estimator. Most LiDAR-aided systems use scan-to-map registration or a LiDAR odometry thread~\cite{zhang2014loam,shan2020lio,xu2022fast,bai2022faster,he2023point,zheng2024fast}. Some continuous-time systems expose raw or point-level LiDAR residuals to a bounded optimization window~\cite{nguyen2023slict,nguyen2024eigen,lang2023coco}. These methods motivate direct LiDAR factors for asynchronous acquisition and motion distortion, but are usually limited to LIO/LIC settings or dedicated continuous-time trajectory parameterizations.

Many multi-sensor systems remain tied to fixed sensor stacks or coupled subsystems~\cite{yang2019dre,10.1109/ICRA48506.2021.9561996,yin2024ground}. Changing the deployment configuration may affect the state definition, factor activation, calibration handling, and marginalization interface. \ultrafusion addresses this gap with a \emph{Unified Sliding-Window Estimator}, where LiDAR residuals, visual reprojection factors, inertial constraints, wheel factors, and GNSS anchors share one state, reliability scheduler, and initialization strategy. This retains direct LiDAR factors while supporting WIO, VIO, LIO, LVIO and their variants in one estimator. Table~\ref{slam_comparison} summarizes representative fusion systems.

\subsection{Sensor Degradation and Spatiotemporal Miscalibration}

Sensor degradation is a common source of SLAM performance loss. Illumination variation, motion blur, dynamic occlusion, geometric degeneracy, wheel slip, or satellite blockage reduce reliable observations and increase outliers; without reliability control, they may cause tracking loss or drift. Degeneracy-aware SLAM methods adapt the estimator through gating, down-weighting, or modality switching~\cite{han2023dams,yin2024ground,yin2023sky}. Ground-Fusion~\cite{yin2024ground} and Ground-Fusion++~\cite{zhang2025towards} demonstrate reliability-conditioned fusion~\cite{jiang2024innovation}, but remain tied to limited modalities or subsystem-level decisions. \ultrafusion applies reliability control directly to LiDAR, vision, IMU, wheel, and GNSS factors inside the common optimization problem.

\emph{Spatiotemporal miscalibration} is another source of bias. Time offsets may arise from asynchronous triggering, latency, or imperfect timestamping, while extrinsics may change with vibration, temperature, or reconfiguration; in large-scale ITS platforms, small errors can bias cross-modal alignment and accumulate drift. Prior work studies joint spatial--temporal calibration~\cite{furgale2013unified,heng2013camodocal} and calibration toolboxes~\cite{koide2023singleshot}, but these methods are usually offline or pre-deployment. \ultrafusion refines time-offset and extrinsic variables inside the SLAM loop when observability and sensor reliability are sufficient.

\subsection{SLAM Benchmark Datasets}

Benchmark datasets are essential for reproducible SLAM evaluation. For ITS-oriented research, an informative benchmark should cover transportation scenarios, diverse sensor configurations, degradation cases, and trustworthy ground truth. Existing datasets often cover only part of this scope: some focus on limited modalities, some provide mild environmental variation, and others rely on proxy trajectories that may be unreliable in difficult sequences.
For example, EuRoC and TUM-VI mainly support visual--inertial evaluation in relatively controlled environments~\cite{burri2016euroc,schubert2018tum}, while datasets such as OpenLORIS-Scene and Ground-Challenge broaden the sensing setup but still have limited coverage of systematic degradation conditions~\cite{shi2020we,yin2023ground}. More recent datasets expand modality and scenario diversity for road, aerial, and non-road mobility~\cite{yin2021m2dgr,yin2024ground,jeong2019complex,li2024mars,frey_tuna2026grandtour}, yet large-scale stress testing under controlled degradation and calibration perturbation remains relatively underexplored.

This motivates benchmarks with rich sensing streams and systematic robustness evaluation~\cite{xu2022review,he2025ligo,yin2023sky,hua2023m2c,yin2024ground}. We present \mthreeDGR, a sensor-rich benchmark for staged degradation, including visibility challenges, geometric degeneracy, wheel slip, and GNSS denial~\cite{zhang2025towards}. \mthreeDGR provides synchronized and calibrated sensor streams for analyzing degradation robustness and spatiotemporal consistency. On this benchmark, we evaluate \textbf{over 60} representative SLAM systems and analyze common failure modes.

\begin{figure*}[htbp]
    \centering
    \includegraphics[width=1.8\columnwidth]{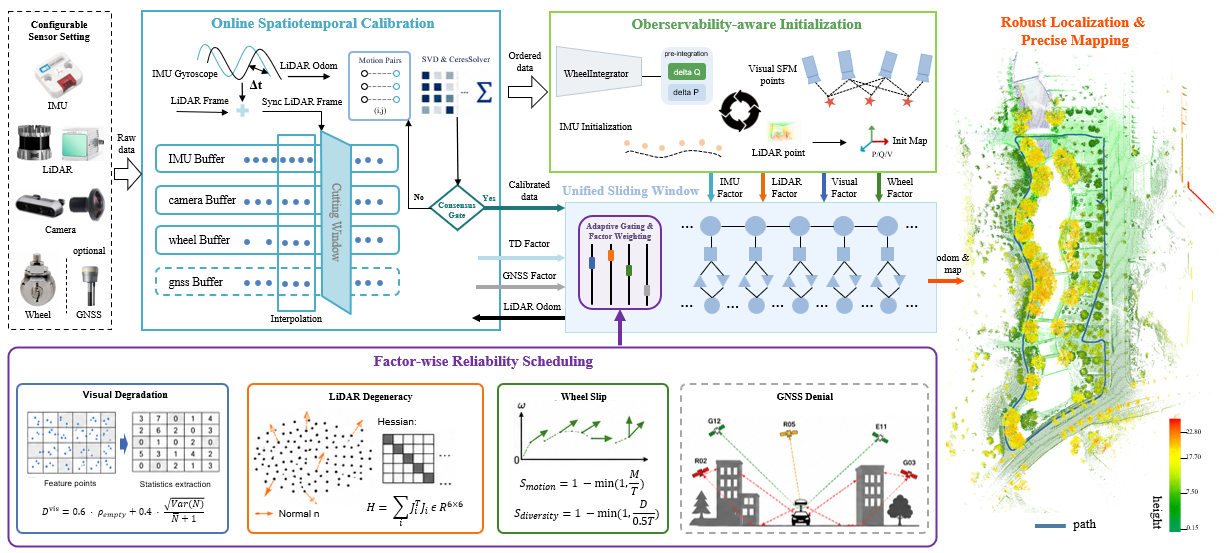}
    \caption{Overview of the \ultrafusion framework. Configurable sensor streams are timestamp-ordered, calibrated online, and initialized under observability-aware bootstrap before entering a unified sliding-window factor graph. Factor-wise reliability scheduling suppresses degraded measurements, while retained factors are jointly optimized across WIO, VIO, LIO, and LVIO configurations with shared state, marginalization, optional wheel/GNSS augmentation, and online calibration.} 
    \label{fig:pipeline}
    \vspace{-7mm}
\end{figure*}


\section{Methodology}
Building on our IROS 2025 Ground-Fusion++ system~\cite{zhang2025towards}, \ultrafusion reorganizes the conference baseline into a journal-ready ITS localization framework through five extensions: (i)~a unified optional-factor sliding-window estimator rather than subsystem coupling, (ii)~observability-aware initialization, (iii)~in-graph factor-wise reliability scheduling, (iv)~online LiDAR--IMU spatiotemporal calibration, and (v)~an expanded \mthreeDGR benchmark with simulation perturbations and cross-platform validation on KAIST, GrandTour, and MARS-LVIG.

Unlike scan-to-map LIO/LVIO pipelines that inject LiDAR odometry as an external prior~\cite{shan2020lio,xu2022fast,bai2022faster,zheng2024fast}, and unlike Ground-Fusion++ where reliability and calibration remain largely subsystem-specific, \ultrafusion keeps LiDAR geometric residuals together with visual, inertial, wheel, and GNSS factors in one shared sliding window. Compared with full continuous-time LIO/LIC systems~\cite{dellenbach2022ct,nguyen2023slict,nguyen2024eigen,lang2023coco}, this retains point-level LiDAR constraints while preserving a compact optimization structure.
As shown in Fig.~\ref{fig:pipeline}, timestamp-ordered measurements pass through online calibration and observability-aware bootstrap before entering the factor graph, where reliability scheduling suppresses unreliable modalities and the active window jointly optimizes retained factors. The following subsections present state representation, the unified estimator, initialization, reliability scheduling, online calibration, and mapping-oriented extensions.



\vspace{-0.5mm}
\subsection{State Representation and Temporal Ordering}
At timestamp $k$, the platform state is
\begin{equation}\scriptsize 
\bm{x}_k \triangleq \big\{\mathbf{R}_k,\ \bm{t}_k,\ \bm{v}_k,\ \bm{b}_{a,k},\ \bm{b}_{g,k}\big\},
\end{equation}
where $\mathbf{R}_k\!\in\!SO(3)$, $\bm{t}_k,\bm{v}_k\!\in\!\mathbb{R}^3$, and $\bm{b}_{a,k},\bm{b}_{g,k}\!\in\!\mathbb{R}^3$. The active window $\mathcal{X}\!=\!\{\bm{x}_{k-W+1},\dots,\bm{x}_k\}$ receives timestamp-ordered constraints from all available sensors. LiDAR extrinsics are $\mathbf{T}_{IL}\!=\!\{\mathbf{R}_{IL},\bm{t}_{IL}\}$, while camera/wheel extrinsics and temporal offsets are
\begin{equation}\scriptsize
\mathbf{T}_{IC}=\{\mathbf{R}_{IC},\bm{t}_{IC}\},\quad
\mathbf{T}_{IO}=\{\mathbf{R}_{IO},\bm{t}_{IO}\},\quad
t_d^{C},\quad t_d^{O},\quad \Delta t_{LI},
\end{equation}
where $t_d^{C}$, $t_d^{O}$, and $\Delta t_{LI}$ denote camera--IMU, wheel--IMU, and LiDAR--IMU time offsets, respectively.

Although the estimator is keyframe-indexed, intra-frame measurements query a local continuous-time segment. For a LiDAR point acquired at normalized time $\alpha_i\in[0,1]$, the deskewing pose is
\begin{equation}\scriptsize
\mathbf{R}_{k}(\alpha_i)=\mathrm{Slerp}\!\left(\mathbf{R}_{k}^{b},\mathbf{R}_{k}^{e};\alpha_i\right),\quad
\bm{t}_{k}(\alpha_i)=(1-\alpha_i)\bm{t}_{k}^{b}+\alpha_i\bm{t}_{k}^{e},
\end{equation}
where superscripts $b$ and $e$ denote scan begin and end poses. This preserves point-time coupling~\cite{dellenbach2022ct,nguyen2023slict,nguyen2024eigen} without introducing a separate batch trajectory.
Pose variables are updated on the manifold using a right-multiplicative local parameterization:
\begin{equation}\scriptsize 
\begin{aligned}
\mathbf{R}_k \leftarrow \mathbf{R}_k\,\mathrm{Exp}\!\big([\delta\bm{\theta}_k]_\times\big),
\\
\bm{t}_k \leftarrow \bm{t}_k + \delta\bm{t}_k,
\end{aligned}
\end{equation}
with $[\cdot]_\times$ the skew operator and $\mathrm{Exp}:\mathbb{R}^3\!\rightarrow\!SO(3)$.

\vspace{-0.5mm}
\subsection{Unified Sliding-Window Estimator}

\ultrafusion represents heterogeneous measurements as optional factors sharing one state, prior, calibration parameterization, and reliability scheduler. The core LiDAR--IMU objective is
\begin{equation}\scriptsize
\begin{aligned}
\min_{\mathcal{X}}\quad
&\mathcal{L}_{\mathrm{prior}} \\
&+\sum_{k'\in\mathcal{W}}\!\Big(
\mathcal{L}_{\mathrm{lidar}}(k')
+\lambda_{\mathrm{imu}}\,\mathcal{L}_{\mathrm{imu}}(k')
\Big),
\end{aligned}
\label{eq:total_objective}
\end{equation}
where $\mathcal{W}$ is the active window index set, $\lambda_{\mathrm{imu}}\!\in\![0,1]$ balances inertial coupling, $\rho(\cdot)$ denotes a robust loss, and $\bm{\Omega}_{(\cdot)}$ denotes the information matrix of the corresponding factor.

Available visual, LiDAR, wheel, and GNSS measurements instantiate additional factors, enabling WIO, VIO, LIO, LVIO, and augmented variants without changing the state or marginalization interface. Unlike subsystem-level coordination~\cite{zhang2025towards}, robustness is controlled at factor level through activation, suppression, or down-weighting.

\noindent\textbf{LiDAR geometric factor.}
LiDAR supplies geometric constraints in large-scale transportation scenes. Unlike scan-to-map pipelines that use point-to-plane registration to produce a LiDAR odometry update outside the multi-sensor window~\cite{zhang2014loam,shan2020lio,xu2022fast,bai2022faster}, \ultrafusion keeps the point-to-plane residual inside the shared objective. For a surface point $\bm{p}_i^L$ acquired at intra-scan time $\alpha_i$, the transformed point $\bm{p}_i^w$ is constrained by its matched local plane $\{\bm{n}_i,\bm{s}_i\}$:
\begin{equation}\scriptsize
\bm{p}_i^w = \mathbf{R}_k(\alpha_i)(\mathbf{R}_{IL}\bm{p}_i^L+\bm{t}_{IL})+\bm{t}_k(\alpha_i),
\quad
r_i^{\mathrm{lidar}}=\bm{n}_i^\top(\bm{p}_i^w-\bm{s}_i).
\end{equation}
The corresponding robust factor is
\begin{equation}\scriptsize
\mathcal{L}_{\mathrm{lidar}}=
\sum_{i\in\mathcal{S}_k}\rho\!\left(\omega_i(r_i^{\mathrm{lidar}})^2\right).
\end{equation}
Here $\mathcal{S}_k$ is the accepted surface-feature set and $\omega_i$ encodes local plane uncertainty. Map-neighborhood queries reuse efficient scan-to-map geometry~\cite{xu2022fast,bai2022faster,he2023point}, while optional intensity consistency provides auxiliary constraints in geometrically weak regions when sufficient support exists.

\noindent\textbf{IMU preintegration factor.}
To preserve short-term motion consistency between geometric updates, \ultrafusion keeps a bias-corrected inertial bridge between adjacent window states~\cite{forster2016manifold,qin2018vins}. We denote the compact preintegration residual by $\bm{r}_{\mathrm{imu}}$ and write
\begin{equation}\scriptsize
\mathcal{L}_{\mathrm{imu}}
:= \rho\!\left(
\left\|
\bm{\Omega}_{\mathrm{imu}}^{1/2}
\bm{r}_{\mathrm{imu}}(\bm{x}_{k-1},\bm{x}_{k},\bm{b}_{a},\bm{b}_{g})
\right\|^2
\right).
\end{equation}
The contribution of this factor is controlled by the IMU reliability score and the coupling weight $\lambda_{\mathrm{imu}}$.

\noindent\textbf{Wheel preintegration factor.}
Wheel odometry provides planar motion constraints but is sensitive to slip and kinematic mismatch. We model it as a preintegrated relative-motion factor with IMU-to-wheel extrinsic $\mathbf{T}_{IO}$, wheel scale $\bm{s}$, and temporal compensation~\cite{liu2019visual,Tingda2022VIW}:
\begin{equation}\scriptsize   
\mathcal{L}_{\mathrm{wheel}}
:= \rho\!\left(\left\|\bm{\Omega}_{\mathrm{wheel}}^{1/2}
\bm{r}_{\mathrm{wheel}}(\bm{x}_{k-1},\bm{x}_{k},\mathbf{T}_{IO},\bm{s},t_d^{O})
\right\|^2\right).
\end{equation}
Its influence is governed by wheel-slip consistency, avoiding over-constraint of weakly observable directions.

\noindent\textbf{Visual reprojection factor.}
Visual measurements enter through a temporally compensated reprojection residual using $\mathbf{T}_{IC}$, inverse depth $\lambda_{k-1}$, and offset $t_d^{C}$~\cite{qin2018vins,lin2021r3live,zheng2024fast}:
\begin{equation}\scriptsize
\bm{r}^{\mathrm{vis}}
:=
\pi\!\left(
\mathbf{T}_{IC}^{-1}\mathbf{T}_{k}^{-1}\mathbf{T}_{k-1}\mathbf{T}_{IC}
\left(\frac{1}{\lambda_{k-1}}\tilde{\bm{u}}_{k-1,t_d^{C}}\right)
\right)-\tilde{\bm{u}}_{k,t_d^{C}},
\end{equation}
\begin{equation}\scriptsize   
\mathcal{L}_{\mathrm{vis}} = \rho\!\left(\|\bm{\Omega}_{\mathrm{vis}}^{1/2} \bm{r}^{\mathrm{vis}}\|^2\right).  
\end{equation}
Here $\mathbf{T}_{k-1}$ and $\mathbf{T}_{k}$ are platform poses, $\tilde{\bm{u}}_{\ell,t_d^{C}}$ is a time-compensated normalized image point, and $\pi(\cdot)$ is the normalized-plane projection. Track age, spatial distribution, KLT consistency, and epipolar checks provide reliability evidence for scheduling.

\noindent\textbf{GNSS position anchoring factor.}
When available, GNSS provides integrity-checked global constraints for drift suppression~\cite{cao2022gvins,hua2023m2c,he2025ligo}. For compactness, we first write the optional position anchoring factor:
\begin{equation}\scriptsize
\mathcal{L}_{\mathrm{gnss}}
:= \rho\!\left(
\left\|\bm{\Omega}_{\mathrm{gnss}}^{1/2}
\big(\bm{t}_k-\bm{p}^{\mathrm{gnss}}_k\big)
\right\|^2
\right),
\end{equation}
where $\bm{p}^{\mathrm{gnss}}_k$ is the GNSS position measurement expressed in the estimator frame, and the measurement covariance and factor activation are governed by the GNSS integrity checks in the reliability scheduler. Raw pseudorange/Doppler measurements are handled by separate factors with receiver clock bias/drift, an ECEF anchor, and ENU--local yaw alignment, and are admitted by the same integrity gates.

After assembling active geometric, inertial, calibration, and reliability-gated factors, \ultrafusion solves a robust nonlinear least-squares problem. Historical information is retained by a Gaussian marginalization prior,
\begin{equation}\scriptsize
\mathcal{L}_{\mathrm{prior}}
:= \left\|\bm{\Omega}_{\mathrm{prior}}^{1/2}\bm{r}_{\mathrm{prior}}(\mathcal{X})\right\|^2,
\end{equation}
where $\bm{r}_{\mathrm{prior}}$ is obtained by Schur-complement or QR marginalization~\cite{qin2018vins,von2022dm}. The prior remains compatible across sensor configurations, while eigenvalue truncation and conservative fallback settings improve numerical stability when active factors change.

\vspace{-0.5mm}
\subsection{Observability-Aware Initialization}

A well-constrained initial state is required before activating the unified sliding-window estimator. We formulate initialization as observability-aware model selection: motion excitation and sensing geometry determine whether the estimator uses SfM-based visual--inertial alignment, stationary or wheel-aided inertial alignment, or LiDAR-odometry-aided short-window MAP estimation; otherwise, the initialization window is extended until sufficient evidence is available.

The resulting bootstrap mode is represented by
\begin{equation}\scriptsize
\varrho_{\mathrm{boot}}\in\{\mathsf{D},\mathsf{S},\mathsf{M},\mathsf{A}\},
\end{equation}
where $\mathsf{D}$ denotes the dynamic visual--inertial hypothesis, $\mathsf{S}$ denotes the stationary or wheel-aided inertial hypothesis, $\mathsf{M}$ denotes the LiDAR-odometry-aided MAP hypothesis, and $\mathsf{A}$ denotes deferred initialization under insufficient observability.

\begin{algorithm}[t]
\caption{Observability-Aware Initialization}
\label{alg:obs_init}
\begin{algorithmic}[1]
\REQUIRE Initial buffer $\mathcal{W}_{\mathrm{init}}$, IMU segment $\mathcal{I}_0$, thresholds $\{\tau_{\omega},\tau_v,\tau_p,\tau_{\omega}^{\Sigma},\tau_{a}^{\Sigma},N_{\min}^{\mathrm{feat}},N_{\min}^{\mathrm{lidar}}\}$
\ENSURE Initial ESKF state $\bm{x}^{\mathrm{eskf}}_0$, bootstrap mode $\varrho_{\mathrm{boot}}$, and admission indicator $\pi_{\mathrm{boot}}\in\{0,1\}$
\STATE Compute IMU statistics $(\bar{\bm{a}},\bar{\bm{\omega}},\bm{\Sigma}_a,\bm{\Sigma}_{\omega})$ and gravity-aligned attitude seed $\mathbf{R}_0$
\STATE Evaluate excitation, visual, wheel, and LiDAR-geometry indicators $(E_{\omega},E_v,\bar{N}^{\mathrm{feat}},\bar{p},N^{\mathrm{lidar}})$ on $\mathcal{W}_{\mathrm{init}}$
\IF {$E_{\omega}>\tau_{\omega}$ \AND $\bar{N}^{\mathrm{feat}}\ge N_{\min}^{\mathrm{feat}}$ \AND $\bar{p}\ge\tau_p$}
    \STATE $\varrho_{\mathrm{boot}}\leftarrow\mathsf{D}$ (SfM-based visual--inertial branch)
    \STATE Run SfM + visual--inertial alignment to recover $\{s,\bm{g},\bm{v}_{0:W},\bm{b}_g\}$
    \STATE Repropagate IMU preintegration with updated bias
\ELSIF {$E_{\omega}\le\tau_{\omega}$ \AND $E_v\le\tau_v$ \AND $\|\bm{\Sigma}_{\omega}\|_F\le\tau_{\omega}^{\Sigma}$ \AND $\|\bm{\Sigma}_a\|_F\le\tau_{a}^{\Sigma}$}
    \STATE $\varrho_{\mathrm{boot}}\leftarrow\mathsf{S}$ (stationary/wheel-aided inertial branch)
    \STATE Set inertial seed: $\bm{b}_g^{0}\!\leftarrow\!\bar{\bm{\omega}}$, $\bm{g}^{0}\!\leftarrow\!-g_n\bar{\bm{a}}/\|\bar{\bm{a}}\|$, $\bm{b}_a^{0}\!\leftarrow\!\bar{\bm{a}}+\mathbf{R}_0^\top\bm{g}^{0}$
\ELSIF {$N^{\mathrm{lidar}}\ge N_{\min}^{\mathrm{lidar}}$ \AND LiDAR geometry check passes}
    \STATE $\varrho_{\mathrm{boot}}\leftarrow\mathsf{M}$ (LiDAR-odometry-aided MAP branch)
    \STATE Use scan-matching poses as geometric priors and solve a short-window MAP problem for $\{\bm{v}_{0:W},\bm{b}_a^{0},\bm{b}_g^{0}\}$
    \STATE Repropagate IMU/wheel preintegration with updated bias
\ELSE
    \STATE $\varrho_{\mathrm{boot}}\leftarrow\mathsf{A}$ (deferred initialization)
    \STATE Continue data accumulation and return with $\pi_{\mathrm{boot}}=0$
\ENDIF
\IF {$\varrho_{\mathrm{boot}}\in\{\mathsf{D},\mathsf{S},\mathsf{M}\}$ \AND MCC consistency check passes}
    \STATE Initialize $\bm{x}^{\mathrm{eskf}}_0=\{\mathbf{R}_0,\bm{t}_0,\bm{v}_0,\bm{b}_a^{0},\bm{b}_g^{0},\bm{g}^{0}\}$ and set $\pi_{\mathrm{boot}}=1$
    \STATE Activate LiDAR geometric factors in the unified window
\ELSE
    \STATE Set $\pi_{\mathrm{boot}}=0$ and continue accumulation
\ENDIF
\end{algorithmic}
\end{algorithm}

In Algorithm~\ref{alg:obs_init}, $E_{\omega}$ and $E_v$ quantify motion excitation, $\bar{N}^{\mathrm{feat}}$ and $\bar{p}$ quantify visual support, and $N^{\mathrm{lidar}}$ measures valid LiDAR geometric support. The bootstrap mode $\varrho_{\mathrm{boot}}$ denotes the selected hypothesis, while $\pi_{\mathrm{boot}}=1$ admits initialization after the MCC consistency check. Under the LiDAR-odometry-aided hypothesis ($\mathsf{M}$), gravity-aligned scan-matching poses provide geometric priors for a short-window MAP estimate of velocity and IMU biases.

\vspace{-0.5mm}
\subsection{Factor-Wise Reliability Scheduling}
Reliability scheduling prevents degraded residuals from dominating the optimizer. At each keyframe, modality-specific degeneracy scores $D_k^{(m)}$ are computed from the evidence below, mapped to binary activation variables $s_k^{(m)}$ and, when appropriate, covariance inflation, smoothed with short-horizon hysteresis to avoid frequent switching, and incorporated into the unified sliding-window objective without altering the state definition or marginalization interface.

At time $k$, each modality $m\in\{\mathrm{LiDAR},\mathrm{Visual},\mathrm{IMU},\mathrm{Wheel},\mathrm{GNSS}\}$ is assigned a normalized degeneracy score $D_k^{(m)}\in[0,1]$ and an activation indicator
\begin{equation}\scriptsize
s_k^{(m)}=\mathbf{1}\!\left[D_k^{(m)}\le\tau^{(m)}\ \land\ N_k^{(m)}\ge N_{\min}^{(m)}\right],
\end{equation}
where $N_k^{(m)}$ is the valid observation count. The scheduled objective in the sliding window is
\begin{equation}\scriptsize
\begin{aligned}
\min_{\mathcal{X}}\quad
\mathcal{L}_{\mathrm{prior}}
 +\sum_{k'\in\mathcal{W}}\Big(
&s^{(\mathrm{LiDAR})}_{k'}\mathcal{L}_{\mathrm{lidar}}(k')
+\lambda_{\mathrm{imu}}\,s^{(\mathrm{IMU})}_{k'}\mathcal{L}_{\mathrm{imu}}(k') \\
&+s^{(\mathrm{Visual})}_{k'}\mathcal{L}_{\mathrm{vis}}(k')
+s^{(\mathrm{Wheel})}_{k'}\mathcal{L}_{\mathrm{wheel}}(k')
+s^{(\mathrm{GNSS})}_{k'}\mathcal{L}_{\mathrm{gnss}}(k')
\Big).
\end{aligned}
\end{equation}

All scores use the convention that larger values indicate lower reliability, with modality-wise weights normalized. For non-LiDAR modalities, compact consistency scores are
\begin{equation}\scriptsize
\begin{aligned}
D_k^{(\mathrm{Visual})}
&=w_f\!\left(1-\min(1,N_f/N_f^{\mathrm{ref}})\right)
+w_g(1-G_f)+w_r\min(1,\bar{e}_{\mathrm{repr}}/\tau_r),\\
D_k^{(\mathrm{IMU})}
&=w_e(1-\eta_k^{\mathrm{exc}})
+w_p\min(1,\|\bm{r}_{\mathrm{imu}}\|_{\bm{\Omega}_{\mathrm{imu}}}^{2}/\tau_{\mathrm{imu}})
+w_s\mathbf{1}[\mathrm{saturation}],\\
D_k^{(\mathrm{Wheel})}
&=w_v\min(1,\|\Delta\bm{p}^{w}-\Delta\bm{p}^{I}\|/\tau_v)
+w_{\psi}\min(1,|\Delta\psi^{w}-\Delta\psi^{I}|/\tau_{\psi}),\\
D_k^{(\mathrm{GNSS})}
&=w_q(1-q_{\mathrm{fix}})
+w_{\Sigma}\min(1,\mathrm{tr}(\bm{\Sigma}_{g})/\tau_{\Sigma})
+w_{\nu}\min(1,\|\bm{\nu}_{g}\|_{\bm{\Sigma}_{g}^{-1}}^{2}/\tau_{\nu}),
\end{aligned}
\label{eq:non_lidar_reliability}
\end{equation}
where $N_f$, $G_f$, and $\bar{e}_{\mathrm{repr}}$ denote visual support, $\Delta(\cdot)^w$ and $\Delta(\cdot)^I$ denote wheel- and IMU-predicted increments, and $q_{\mathrm{fix}}$, $\bm{\Sigma}_g$, and $\bm{\nu}_g$ denote GNSS integrity evidence. These cues determine factor admission or covariance inflation within the same scheduler.

\noindent\textbf{LiDAR degeneracy detection.}
LiDAR reliability is evaluated from the point-to-plane Hessian accumulated over the local scan~\cite{han2023dams,yin2024ground}. With Jacobian rows $\bm{J}_i=[\bm{n}_i^\top, -\bm{n}_i^\top[\bm{p}_i]_\times]$,
\begin{equation}\scriptsize
\mathbf{H}_k = \sum_i \bm{J}_i^\top\bm{J}_i + 10^{-8}\mathbf{I}_6,
\end{equation}
Eigen-decomposition yields $\lambda_1\le\cdots\le\lambda_6$ and $\kappa(\mathbf{H}_k)=\lambda_6/(\lambda_1+10^{-12})$. Together with normal covariance $\mathbf{C}_n$ and match count $M_k$, they define
\begin{equation}\scriptsize
\begin{aligned}
D_k^{(\mathrm{LiDAR})} &= w_h\phi_h(\lambda_1,\kappa) + w_n\frac{\tau_n}{\tau_n+\lambda_3(\mathbf{C}_n)} + w_a\phi_a(\mathbf{H}_k) + w_c\Bigl(1-\min\bigl(1,\tfrac{M_k}{M_{\mathrm{ref}}}\bigr)\Bigr),\\
\phi_h(\lambda_1,\kappa) &= \tfrac{1}{2}\!\left[\min\bigl(1,\tfrac{\tau_\lambda}{\lambda_1}\bigr) + \min\bigl(1,\tfrac{\kappa}{\tau_\kappa}\bigr)\right],
\end{aligned}
\end{equation}
where $\phi_a(\cdot)$ penalizes axes with weak constraint projection, $\lambda_3(\mathbf{C}_n)$ is the smallest normal-covariance eigenvalue, $\tau_n,\tau_\lambda,\tau_\kappa$ are thresholds, $M_{\mathrm{ref}}$ is a reference match count, and $w_h,w_n,w_a,w_c$ are normalized non-negative weights. When this score exceeds the modality-specific threshold, the corresponding LiDAR factors are either deactivated through $s_k^{(\mathrm{LiDAR})}$ or attenuated, thereby preventing ill-conditioned scan matches from degrading the joint solution while retaining informative measurements in well-structured regions.

\noindent\textbf{Visual reliability check.}
Visual tracking quality varies sharply with illumination and texture. The scheduler combines feature count $N_f$, spatial distribution uniformity $G_f$ (measured by grid occupancy variance on an $8\times 8$ image partition), forward-backward KLT inlier ratio, and mean reprojection residual $\bar{e}_{\mathrm{repr}}$:
\begin{equation}\scriptsize
D_k^{(\mathrm{Visual})} = w_f\Bigl(1-\min\bigl(1,\tfrac{N_f}{N_f^{\mathrm{ref}}}\bigr)\Bigr) + w_g(1-G_f) + w_r\min\bigl(1,\tfrac{\bar{e}_{\mathrm{repr}}}{\tau_r}\bigr),
\end{equation}
where $N_f^{\mathrm{ref}}$ is a reference track count, $\tau_r$ a residual threshold, and $w_f,w_g,w_r$ are non-negative weights. Unlike systems that mainly reject measurements before optimization~\cite{qin2018vins,lin2021r3live,zheng2024fast}, \ultrafusion incorporates this evidence into factor scheduling. When $D_k^{(\mathrm{Visual})}$ exceeds its threshold or $N_f<N_{\min}^{(\mathrm{Visual})}$, visual factors are deactivated or attenuated.

\noindent\textbf{IMU excitation consistency.}
Inertial preintegration supplies short-term motion continuity but can introduce bias under insufficient excitation~\cite{forster2016manifold,qin2018vins}. The scheduler monitors rotational and translational excitation $\eta_k^{\mathrm{exc}}$, preintegration residual magnitude $\|\bm{r}_{\mathrm{imu}}\|_{\bm{\Omega}_{\mathrm{imu}}}^2$, and saturation flags:
\begin{equation}\scriptsize
D_k^{(\mathrm{IMU})} = w_e(1-\eta_k^{\mathrm{exc}}) + w_p\min\Bigl(1,\tfrac{\|\bm{r}_{\mathrm{imu}}\|_{\bm{\Omega}_{\mathrm{imu}}}^2}{\tau_{\mathrm{imu}}}\Bigr) + w_s\mathbf{1}[\mathrm{saturation}],
\end{equation}
where $\tau_{\mathrm{imu}}$ is a residual threshold and $w_e,w_p,w_s$ weight the terms. When excitation drops below $\tau_\omega$ or residuals spike, the soft weight $\lambda_{\mathrm{imu}} s_k^{(\mathrm{IMU})}$ reduces the influence of unreliable inertial constraints.

\noindent\textbf{Wheel slip consistency.}
Wheel odometry provides high-rate velocity measurements but is susceptible to slip and kinematic model violation~\cite{liu2019visual,Tingda2022VIW}. The scheduler compares wheel-derived increments $(\Delta\bm{p}^w,\Delta\psi^w)$ against inertial and visual predictions $(\Delta\bm{p}^I,\Delta\psi^I)$ within a short temporal window:
\begin{equation}\scriptsize
D_k^{(\mathrm{Wheel})} = w_v\min\Bigl(1,\tfrac{\|\Delta\bm{p}^{w}-\Delta\bm{p}^{I}\|}{\tau_v}\Bigr) + w_{\psi}\min\Bigl(1,\tfrac{|\Delta\psi^{w}-\Delta\psi^{I}|}{\tau_{\psi}}\Bigr),
\end{equation}
where $\tau_v,\tau_{\psi}$ are translational and rotational deviation thresholds, and $w_v,w_{\psi}$ weight the terms. Large discrepancies or low motion diversity inflate the wheel-factor covariance or set $s_k^{(\mathrm{Wheel})}=0$, limiting slip-induced drift while retaining reliable wheel constraints.

\noindent\textbf{GNSS integrity checks.}
GNSS measurements suppress drift but can be corrupted by multipath or satellite blockage~\cite{cao2022gvins,hua2023m2c,he2025ligo}. The scheduler examines fix quality $q_{\mathrm{fix}}$, covariance trace $\mathrm{tr}(\bm{\Sigma}_g)$, and innovation consistency against the local prediction $\|\bm{\nu}_g\|_{\bm{\Sigma}_g^{-1}}^2$:
\begin{equation}\scriptsize
D_k^{(\mathrm{GNSS})} = w_q(1-q_{\mathrm{fix}}) + w_{\Sigma}\min\Bigl(1,\tfrac{\mathrm{tr}(\bm{\Sigma}_g)}{\tau_{\Sigma}}\Bigr) + w_{\nu}\min\Bigl(1,\tfrac{\|\bm{\nu}_g\|_{\bm{\Sigma}_g^{-1}}^2}{\tau_{\nu}}\Bigr),
\end{equation}
where $\tau_{\Sigma},\tau_{\nu}$ are covariance and innovation thresholds, and $w_q,w_{\Sigma},w_{\nu}$ weight the terms. Failed integrity indicators exclude the GNSS factor through $s_k^{(\mathrm{GNSS})}$, protecting the locally consistent trajectory from erroneous global updates.
Thus, reliability control remains within a single optimization problem rather than switching among subsystems~\cite{zhang2025towards}.

\vspace{-0.5mm}
\subsection{Online Spatiotemporal Calibration}
Online Spatiotemporal Calibration (OSC) refines the LiDAR--IMU temporal offset and rotation extrinsic online under sufficient excitation. The calibration bundle is $\bm{\vartheta}_{\mathrm{cal}}=\{\Delta t_{LI},\mathbf{R}_{IL}\}$, with $\bm{t}_{IL}$ held fixed. OSC runs two lightweight workers in parallel: a temporal worker estimates $\Delta t_{LI}$ from IMU--LiDAR motion alignment when excitation and turning guards are satisfied, while an extrinsic worker refines $\mathbf{R}_{IL}$ from scan-to-scan and inertial rotation consistency under multi-axis excitation and adequate scan quality. Candidate updates are admitted only after residual, confidence, and short-history consensus checks; accepted values are then injected into timestamp association, preintegration, and LiDAR deskewing while preserving the current world LiDAR pose.

\vspace{-0.5mm}
\subsubsection{Temporal Calibration}
The temporal worker estimates $\Delta t_{LI}$ on long asynchronous windows using frontend LiDAR odometry. LiDAR-frame angular velocity and LiDAR-derived acceleration surrogates are computed from consecutive front-end LiDAR odometry poses and interpolated using cubic Hermite splines, providing continuous motion cues for temporal alignment with IMU measurements. We use $\delta$ as the LiDAR-trajectory search shift: an IMU sample at $t_i$ is matched to LiDAR motion at $t_i+\delta$, while the runtime offset is defined by
\begin{equation}\scriptsize
t_{\mathrm{imu}} = t_{\mathrm{lidar}}+\Delta t_{LI},
\qquad
\Delta t_{LI}=-\delta.
\end{equation}
A coarse candidate is initialized by maximizing the cross-correlation between IMU and shifted LiDAR motion norms,
\begin{equation}\scriptsize
\hat{\delta}_{\mathrm{coarse}}
=
\arg\max_{\delta\in[-\Delta t_{\max},\Delta t_{\max}]}
\mathcal{C}\!\left(\|\bm{\omega}^{I}(t)\|,\ \|\bm{\omega}^{L}(t+\delta)\|\right),
\end{equation}
optionally augmented by correlation between IMU specific-force norms and LiDAR-derived acceleration surrogate norms. The estimate is refined by minimizing a robust alignment cost on matched pairs,
\begin{equation}\scriptsize
\mathcal{J}_{LI}(\delta)
=
\frac{1}{N}\sum_{i}
\rho_{\mathrm{H}}\!\left(
\left\|
\bm{\omega}^{I}_i
-
\mathbf{R}_{IL}\bm{\omega}^{L}(t_i+\delta)
-
\bm{b}_g
\right\|
\right)
+
\lambda_a
\rho_{\mathrm{H}}\!\left(
\left\|
s\bm{a}^{I}_i
-
\mathbf{R}_{IL}\tilde{\mathbf{a}}^{L}(t_i+\delta)
-
\bm{b}_a
\right\|
\right),
\end{equation}
where $\rho_{\mathrm{H}}$ is a Huber loss and $(\mathbf{R}_{IL},\bm{b}_g,\bm{b}_a,s)$ are estimated jointly when the extrinsic is not yet locked; otherwise $\mathbf{R}_{IL}$ is fixed and only $\delta$ and gyro bias are refined. Accepted candidates must exceed confidence and excitation thresholds, improve the cost relative to a zero-offset baseline, and remain stable over a short candidate history; updates are suppressed during sharp turning. Repeated stable commits freeze $\Delta t_{LI}$ for the runtime pipeline.

\vspace{-0.5mm}
\subsubsection{LiDAR--IMU Extrinsic Calibration}
The extrinsic worker refines $\mathbf{R}_{IL}$ from motion-consistency constraints between inertial preintegration and a dedicated scan-to-scan LiDAR odometry branch that is isolated from the main SLAM frontend. For a motion pair over $[i,j]$, gyro integration gives
\begin{equation}\scriptsize
\mathbf{R}^{I}_{ij}
=
\prod_{k=i}^{j-1}
\mathrm{Exp}\!\big((\bm{\omega}_k-\bm{b}_g)\Delta t_k\big),
\end{equation}
while calibration-branch scan-to-scan registration accumulates $\mathbf{R}^{L}_{ij}$. Under
\begin{equation}\scriptsize
\mathbf{R}^{I}_{ij}=\mathbf{R}_{IL}\mathbf{R}^{L}_{ij}\mathbf{R}_{IL}^{\top},
\end{equation}
the rotation vectors satisfy the linearized constraint
\begin{equation}\scriptsize
\bm{\phi}^{I}_{ij}\approx \mathbf{R}_{IL}\bm{\phi}^{L}_{ij},
\qquad
\bm{\phi}^{I}_{ij}=\mathrm{Log}(\mathbf{R}^{I}_{ij}),
\quad
\bm{\phi}^{L}_{ij}=\mathrm{Log}(\mathbf{R}^{L}_{ij}).
\end{equation}
Each pair is weighted by $w_{ij}$, which increases with scan-to-scan alignment quality. Calibration is admitted only when multi-axis excitation is sufficient: with $\bm{a}_k=\bm{\phi}^{L}_k/\|\bm{\phi}^{L}_k\|$ and $\mathbf{M}=\sum_k w_k\bm{a}_k\bm{a}_k^{\top}$, we require enough valid pairs, adequate accumulated rotation, and a lower bound on $\lambda_1/\lambda_3$ from the eigendecomposition of $\mathbf{M}$ to avoid single-axis degeneracy.

An initial rotation is obtained from weighted angular-velocity alignment,
\begin{equation}\scriptsize
\min_{\mathbf{R}_{IL},\bm{b}_g}
\sum_k
w_k
\left\|
\mathbf{R}_{IL}\bar{\bm{\omega}}^{L}_k+\bm{b}_g-\bar{\bm{\omega}}^{I}_k
\right\|^2,
\end{equation}
which admits a closed-form Procrustes solution via SVD on the centered cross-covariance. If excitation or residual checks fail, a rotation-vector fallback minimizes $\sum_k w_k\|\mathbf{R}_{IL}\bm{\phi}^{L}_k-\bm{\phi}^{I}_k\|^2$ with the same SVD structure. The estimate is further refined by
\begin{equation}\scriptsize
\mathbf{R}_{IL}=\mathrm{Exp}(\delta\bm{\theta})\mathbf{R}_{0},
\qquad
\min_{\delta\bm{\theta},\bm{b}_g}
\sum_k
\left\|e^R_k\right\|^2
+
\sum_k
\left\|e^\omega_k\right\|^2,
\end{equation}
where $e^R_k$ and $e^\omega_k$ are weighted rotation-vector and angular-velocity residuals around $\mathbf{R}_0$. Candidates passing mean/max residual, inlier-ratio, and excitation tests enter a consensus history; $\mathbf{R}_{IL}$ is locked once repeated agreement is reached across successful solves.

To avoid map discontinuity at lock time, the updated extrinsic is applied while preserving the current world LiDAR pose,
\begin{equation}\scriptsize
\mathbf{T}^{new}_{WI}
=
\mathbf{T}^{old}_{WI}\mathbf{T}^{old}_{IL}(\mathbf{T}^{new}_{IL})^{-1},
\qquad
\mathbf{T}^{new}_{IL}=\{\mathbf{R}^{\mathrm{locked}}_{IL},\bm{t}^{old}_{IL}\},
\end{equation}
so that $\mathbf{T}^{new}_{WL}=\mathbf{T}^{old}_{WL}$ and the existing voxel map remains consistent with incoming scans.

\vspace{-1mm}
\subsection{Localization and Mapping Refinements}
Several implementation refinements improve repeatability and mapping quality without changing the estimator:

\noindent\textbf{Robust localization.}
Before residual construction, image/depth timestamps, LiDAR point times, intensity fields, trajectory frames, and ground-truth associations are aligned. Factors are restricted to physically supported state directions; confirmed low-excitation stops use a ZUPT-inspired branch~\cite{gui2023zupt}, wheel factors are validated by IMU--wheel consistency, and LVIO feature frames are synchronized with the LiDAR window before reprojection.

\noindent\textbf{Geometric and colorized mapping.}
State estimation is accompanied by a hybrid local map: a bounded voxel-hash map supports nearest-neighbor search, and a sliding downsampled map provides geometric support and plane-quality evidence~\cite{yuan2022efficient,bai2022faster}. Deskewed scans are transformed into the world frame and colorized through temporally aligned RGB projection for visualization.
The mapping interface also supports LiDAR-guided 3D Gaussian Splatting~\cite{lang2025gaussianlic2}. \ultrafusion poses, aligned RGB images, and colorized LiDAR points provide metric anchors, sparse depth supervision, and appearance constraints for incremental Gaussian construction without replacing the odometry frontend.

\section{\mthreeDGRc Benchmark Dataset}
\mthreeDGR is designed for controlled evaluation of multi-sensor SLAM under degradation, calibration uncertainty, and diverse motion regimes. Its construction emphasizes broad modality coverage, accurate synchronization and calibration, and scenarios targeting representative transportation failure modes.

\noindent\textbf{Relation to prior datasets.}
An earlier IROS version of \mthreeDGR introduced the real-world degradation dataset and evaluated 40 representative SLAM systems~\cite{zhang2025towards}. This paper extends that benchmark in two ways: it adds simulation trajectories for controlled perturbation analysis, especially for LiDAR degeneracy and spatiotemporal miscalibration, and expands the comparison to more than 60 systems. The expanded benchmark is used both as a dataset contribution and as a stress-test protocol for configurable multi-sensor fusion systems.
\vspace{-0.5mm}
\subsection{Real-World Acquisition}
\begin{figure*}[htbp]
    \centering
    \includegraphics[width=1.7\columnwidth]{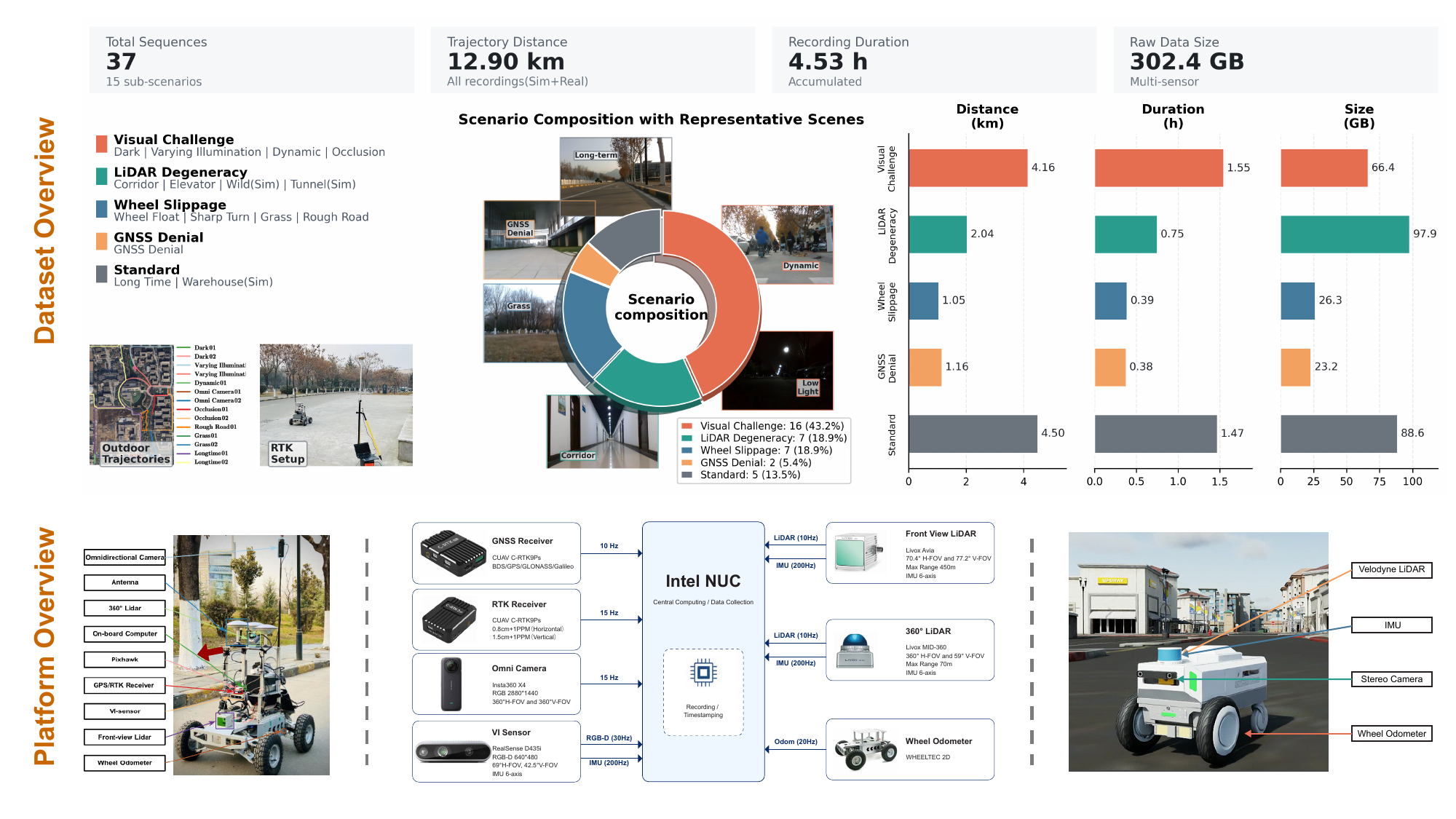}
    \vspace{-4mm}
    \caption{Overview of the \mthreeDGR benchmark and acquisition platforms. The upper panel summarizes the dataset scale, scenario composition, representative outdoor trajectories, and distance--duration--storage statistics across visual challenge, LiDAR degeneracy, wheel slippage, GNSS-denial, standard, and simulation sequences. The lower panel shows the real ground-robot platform and the \mthreeDGR Sim robot, including mechanical layout, sensor placement, and data links among the RGB-D--IMU camera, omnidirectional camera, dual LiDARs, GNSS/RTK receivers, wheel odometer, Pixhawk, onboard computer, and simulated IMU--wheel--LiDAR--stereo setup.} 
    \label{robot_car}
    \vspace{-5mm}
\end{figure*}

Figure~\ref{robot_car} summarizes the \mthreeDGR acquisition setting, including benchmark scale, scenario taxonomy, representative trajectories, and both real and simulated sensing platforms. The real ground robot records RGB-D imagery, LiDAR point clouds, wheel encoder odometry, and raw GNSS measurements. LiDAR streams are transmitted through Ethernet, while the remaining sensors use USB 3.2; all data are logged on an Intel NUC with a high-speed NVMe SSD for long-duration synchronized acquisition.

As shown in the platform panel of Figure~\ref{robot_car}, the robot uses a differential-drive base with two driven wheels, which naturally provides wheel-odometry measurements. Visual sensing is provided by an RGB-D--IMU device and an omnidirectional camera. Two Livox solid-state LiDARs provide complementary 3D observations with non-repetitive scanning patterns, while GNSS/RTK receivers mounted on the top layer record satellite measurements for outdoor evaluation. For reference trajectories, we use an RTK receiver in outdoor sequences and a motion-capture system in indoor sequences.

The real-world part of \mthreeDGR is organized around sensing failures that commonly affect ground mobility: appearance degradation for cameras, weak geometric constraints for LiDAR, unreliable wheel motion under slip or rough terrain, and intermittent GNSS availability. This design keeps the platform close to practical ITS deployment while making each dominant failure source observable during evaluation.

All sensor topics are recorded as ROS bags with a unified timestamping mechanism. In addition, several devices provide internal hardware-level synchronization to further reduce inter-sensor skew. Ground truth is provided by motion capture indoors, RTK GNSS outdoors, and ArUco-based start--end alignment for drift evaluation under degeneracy or GNSS outage, so both absolute trajectory accuracy and accumulated loop drift can be assessed.

\vspace{-0.5mm}
\subsection{\mthreeDGR Sim Acquisition}
The \mthreeDGR Sim trajectories are recorded in NVIDIA Isaac Sim using a differential-drive mobile robot modeled after Nova Carter. As shown in Figure~\ref{robot_car}, the simulated platform provides a compact multi-sensor layout with body-mounted IMU, LiDAR, and stereo camera streams, while wheel-encoder measurements are generated from the two driven wheels.

For each scene, a closed-loop collection pipeline first constructs a 2D occupancy grid from multi-slice LiDAR projections in the 3D environment, then plans collision-free routes with obstacle-inflated A* search, simplifies them with Ramer--Douglas--Peucker, and tracks them using pure pursuit control. During autonomous motion, the simulated IMU, wheel encoder, LiDAR, and stereo camera share the same simulation-time clock, are published through the ROS~1 bridge, and are logged as synchronized bags together with simulator-derived sensor extrinsics as ground-truth calibration.

The simulated sequences complement real data by providing repeatable routes and exact calibration references. They are mainly used for controlled LiDAR-degeneracy and spatiotemporal-perturbation studies, where the same scene and motion can be replayed while changing timing offsets, extrinsic errors, or geometric structure.

\vspace{-0.5mm}
\subsection{Sequence Design}
The sequences include routine operation and deliberately challenging scenes, enabling analysis of both average accuracy and modality-specific failure modes. Real-world sequences cover visual degradation, LiDAR degeneracy, wheel slip, GNSS outage, and standard short- and long-term operation, while \mthreeDGR Sim adds Wild, Warehouse, and Tunnel trajectories for repeatable LiDAR-degeneracy and spatiotemporal-perturbation studies.

Each sequence group targets a different robustness question. Visual challenge sequences test whether localization remains stable under low light, illumination changes, dynamics, or occlusion; LiDAR-degenerate sequences emphasize corridor-like and elevator-transition geometry; wheel-slippage sequences expose biased planar motion constraints; GNSS-denial sequences test continuity without global updates; and standard routes provide nominal short- and long-duration baselines. Figure~\ref{robot_car} visualizes representative trajectories by scenario; detailed sequence definitions, ground-truth sources, and aggregate statistics are provided in the supplementary material.

\section{Experimental Evaluation}

Evaluation is organized around configuration-wise accuracy, module causality, degradation robustness, calibration robustness, long-horizon/high-speed operation, and cross-platform validation.

\begin{itemize}
\item \textbf{Q1. Overall Benchmarking:} Does the \ultrafusion system with diverse configurations (WIO/VIO/LIO/LVIO with optional augmentation when applicable) achieve competitive performance against corresponding baselines?
\item \textbf{Q2. Degradation Robustness:} Does the degradation-aware tightly-coupled design improve robustness under modality-specific failure modes?
\item \textbf{Q3. Spatiotemporal Calibration:} Does online LiDAR--IMU temporal and extrinsic calibration remain effective under injected delay and rotation perturbations?
\item \textbf{Q4. Long-Term and High-Speed Operation:} Does \ultrafusion remain robust and stable over long-duration and high-speed trajectories?
\item \textbf{Q5. Cross-Platform Validation:} Does \ultrafusion remain effective on representative heterogeneous robotic platforms?
\end{itemize}

Experiments use \mthreeDGR~\cite{zhang2025towards} for degradation and controlled perturbation on a wheeled ground robot, M2DGR-Plus~\cite{yin2024ground} for campus-scale wheeled routes, KAIST~\cite{jeong2019complex} for city-scale driving up to 96.9~km/h, GrandTour~\cite{frey_tuna2026grandtour} for quadruped mobility, and MARS-LVIG~\cite{li2024mars} for low-altitude UAV trajectories along airport infrastructure.

\vspace{-0.5mm}
\subsection{Comprehensive Benchmarking}
Table~\ref{rmse_tab} compares \ultrafusion with representative systems on ten \mthreeDGR sequences, emphasizing accuracy and robustness under heterogeneous degradation. M2DGR-Plus results are reported in Table~\ref{time_asynchronous_tab}.

\textbf{Baselines and evaluation metrics.}
Baselines are grouped by compatible sensor availability: wheel/GNSS references, visual systems without LiDAR~\cite{mur2017orb,wang2021tartanvo,murai2024_mast3rslam,campos2021orb,von2022dm,qin2018vins,shan2019rgbd,cao2022gvins,Tingda2022VIW,Wallong2021,yin2024ground}, LiDAR-only or LiDAR--inertial systems~\cite{qin2018aloam,lin2020loam,chengwei2023CTLO,shan2018lego,vizzo2023ral,pan2024tro,Yu2024I2EKFLOAD,zheng2024traj,shen2025cte,10891038,lee2024genzicp,li2021towards,xu2022fast,bai2022faster,he2023point,huang2023log,chengwei2023CT,Livox2021LIO,chen2023direct,qin2020lins,ma2024mm,chengwei2023hmlio,chengwei2022iESKFlio,yuan2022efficient,ye2019tightly,huang2024loglio2lidarinertialodometryefficient,nguyen2023slict,nguyen2024eigen,liu2024voxel,10806842,zou2024ltaom,tao2024eqlio,malladi2025arxiv,10321658,10610818,10966190,choi2025surfelliofastlidarinertialodometry}, and LiDAR--visual systems~\cite{10.1109/ICRA48506.2021.9561996,lin2021r2live,lin2021r3live,lang2023coco,zheng2022fast,zheng2024fast,yuan2024sr,zhang2025towards}. Each method uses its supported streams, and each \ultrafusion variant uses only the modalities indicated by its mode name. Accuracy is reported as EVO-aligned ATE RMSE. A run is marked by \blackx if initialization, tracking, execution, or temporal alignment fails; failed runs are penalized in aggregate ATE as specified in the table note. Repository links and licenses are summarized in the supplementary material.
\textbf{We will release the implementation and evaluation scripts upon paper acceptance.}

\textbf{Overall performance.}
Table~\ref{rmse_tab} reveals modality-dependent failure modes. Visual pipelines degrade under poor correspondence~\cite{campos2021orb,wang2021tartanvo,murai2024_mast3rslam}, whereas LiDAR-based systems tolerate appearance changes but may fail under geometric degeneracy~\cite{xu2022fast,zheng2024fast,yuan2024sr}. Average rank is therefore more informative than isolated best-case accuracy.

\ultrafusion achieves favorable ranks across sensor groups. WIO improves over raw wheel odometry and Ground-Fusion WIO, while VWIO, LWIO, and LVWIO lead the visual-based, LiDAR-only, and LiDAR--visual groups. The gains are consistent with observability-aware initialization and factor-level fusion: wheel/GNSS availability changes active factors without altering the estimator, and LiDAR residuals remain optimized in the same window rather than injected as an external odometry prior. On \mthreeDGR, lower ATE under visual, LiDAR-degenerate, wheel-slip, and GNSS-denial sequences reduces pose errors in tunnel transit, degraded-camera driving, and mixed indoor--outdoor mobility. On M2DGR-Plus (Table~\ref{time_asynchronous_tab}), \ultrafusion achieves the lowest average drift rate and RMSE (0.59\% / 0.24 m), compared with 2.32\% / 1.48 m for FAST-LIVO2~\cite{zheng2024fast} and 1.71\% / 0.75 m for Ground-Fusion~\cite{yin2024ground}.

\begin{table*}[htbp]
    \caption{ATE RMSE (m) comparison of representative SLAM systems on \mthreeDGR sequences.}
    \label{rmse_tab}
    \renewcommand{\arraystretch}{1.3}
    \begin{adjustbox}{width=1.9\columnwidth}
    \centering
    \begin{tabular}{*{16}c}
        \hline
        \multirow{2}{*}{\makecell{{Method/Scenario}}}
          & \multicolumn{1}{c}{\makecell{Summary}}

          & \multicolumn{4}{c}{\makecell{Visual Challenge}}
          &
          & \multicolumn{2}{c}{\makecell{LiDAR Degeneracy}}
          &
          & \multicolumn{3}{c}{\makecell{Wheel Slippage}}
          &
          & \multicolumn{1}{c}{\makecell{GNSS Denial}} \\
        \cline{2-2} \cline{3-6}  \cline{8-9} \cline{11-13} \cline{15-15}
        & Avg. Rank$^1$/ATE& Dynamic01 & Varying-illu01 & Dark01 & Occlusion01 & & Corridor01 & Elevator01 & & Wheel-float01 & Sha-turn01 & Grass01 & & GNSS-denial01\\
        \hline




        
        Raw Wheel Odom 
        & \rd{2.5/35.6} 
        & \rd{8.60} 
        & \rd{8.21} 
        & \fs\bf{5.48} 
        & \rd{6.90} 
        & 
        & \rd{74.32} 
        & \nd\sl{66.94} 
        & 
        & \nd\sl{1.18} 
        & \nd\sl{7.44} 
        & \makecell[c]{26.95} 
        & 
        & \makecell[c]{\blackx}\\
        
        Ground-Fusion WIO\cite{yin2024ground} 
        & \nd\sl{2.0/33.68} 
        & \nd\sl{2.32} 
        & \nd\sl{2.36} 
        & \nd\sl{5.52} 
        & \nd\sl{2.04} 
        & 
        & \nd\sl{72.61} 
        & \nd\sl{66.94} 
        & 
        & \rd{2.20} 
        & \nd\sl{7.44} 
        & \rd{25.34} 
        & 
        & \blackx\\
        
        GNSS SPP 
        & \makecell[c]{2.7/106.98} 
        & \blackx 
        & \blackx 
        & \makecell[c]{7.69} 
        & \blackx 
        & 
        & \blackx 
        & \blackx 
        & 
        & \blackx 
        & \blackx 
        & \fs\bf{0.48} 
        & 
        & \fs\bf{11.61}\\
        
        \bf \ultrafusion (WIO), 2026
        & \fs\bf{1.3/26.99}
        & \fs\bf{0.55}
        & \fs\bf{0.48}
        & \rd{5.59}
        & \fs\bf{0.81}
        &
        & \fs\bf{24.67}
        & \fs\bf{64.55}
        &
        & \fs\bf{0.49}
        & \fs\bf{2.06}
        & \nd\sl{20.65}
        &
        & \makecell[c]{\blackx}\\
        \hline


       ORB-SLAM2\cite{mur2017orb}, 2017
       & \makecell[c]{7/76.79}
       & \fs\bf{0.14}
       & \blackx
       & \blackx
       & \blackx
       &
       & \makecell[c]{6.41}
       & \makecell[c]{8.09}
       &
       & \makecell[c]{1.72}
       & \makecell[c]{1.54}
       & \blackx
       &
       & \blackx \\

        VINS-Mono\cite{qin2018vins}, 2018
        & \makecell[c]{6.4/26.7}
        & \makecell[c]{0.43}
        & \makecell[c]{2.70}
        & \makecell[c]{7.91}
        & \blackx
        &
        & \makecell[c]{9.82}
        & \makecell[c]{62.80}
        &
        & \makecell[c]{0.46}
        & \makecell[c]{0.36}
        & \rd{2.17}
        &
        & \makecell[c]{30.36} \\

       VINS-RGBD\cite{shan2019rgbd}, 2019
       & \makecell[c]{5.7/49.21}
       & \rd{0.20}
       & \makecell[c]{1.86}
       & \blackx
       & \blackx
       &
       & \makecell[c]{5.62}
       & \blackx
       &
       & \makecell[c]{0.28}
       & \rd{0.35}
       & \makecell[c]{7.52}
       &
       & \makecell[c]{26.31} \\

       TartanVO\cite{wang2021tartanvo}, 2021
       & \makecell[c]{8.5/62.56}
       & \makecell[c]{2.37}
       & \makecell[c]{2.17}
       & \makecell[c]{12.37}
       & \blackx
       &
       & \blackx
       & \blackx
       &
       & \makecell[c]{1.93}
       & \makecell[c]{2.09}
       & \makecell[c]{4.68}
       &
       & \blackx \\

       ORB-SLAM3\cite{campos2021orb}, 2021
       & \makecell[c]{9.8/150}
       & \blackx
       & \blackx
       & \blackx
       & \blackx
       &
       & \blackx
       & \blackx
       &
       & \blackx
       & \blackx
       & \blackx
       &
       & \blackx\\

       VINS-GPS-Wheel\cite{Wallong2021}, 2021
       & \makecell[c]{6.7/25.53}
       & \makecell[c]{1.18}
       & \makecell[c]{1.32}
       & \makecell[c]{15.55}
       & \blackx
       &
       & \rd{5.55}
       & \makecell[c]{43.48}
       &
       & \makecell[c]{0.86}
       & \makecell[c]{2.00}
       & \makecell[c]{18.47}
       &
       & \makecell[c]{16.89}\\

       DM-VIO\cite{von2022dm}, 2022
       & \makecell[c]{7.7/63.22}
       & \makecell[c]{2.25}
       & \makecell[c]{2.27}
       & \makecell[c]{4.08}
       & \blackx
       &
       & \makecell[c]{12.20}
       & \fs\bf{2.54}
       &
       & \blackx
       & \makecell[c]{8.90}
       & \blackx
       &
       & \blackx\\

       GVINS\cite{cao2022gvins}, 2022
       & \makecell[c]{5.7/61.45}
       & \makecell[c]{0.26}
       & \makecell[c]{1.25}
       & \blackx
       & \blackx
       &
       & \makecell[c]{9.42}
       & \nd\sl{2.89}
       &
       & \rd{0.27}
       & \makecell[c]{0.40}
       & \blackx
       &
       & \blackx \\

       VIW-Fusion\cite{Tingda2022VIW}, 2022
       & \makecell[c]{5.5/27.99}
       & \makecell[c]{0.62}
       & \makecell[c]{1.02}
       & \fs\bf{0.77}
       & \blackx
       &
       & \makecell[c]{5.58}
       & \makecell[c]{16.68}
       &
       & \makecell[c]{0.77}
       & \makecell[c]{2.44}
       & \makecell[c]{2.91}
       &
       & \makecell[c]{99.06}\\

       Ground-Fusion\cite{yin2024ground}, 2024
       & \rd{4/7.52}
       & \nd\sl{0.19}
       & \nd\sl{0.59}
       & \nd\sl{1.10}
       & \fs\bf{1.21}
       &
       & \makecell[c]{26.25}
       & \makecell[c]{29.93}
       &
       & \makecell[c]{0.29}
       & \makecell[c]{1.16}
       & \nd\sl{1.33}
       &
       & \rd{13.19}\\

       MASt3R-SLAM\cite{murai2024_mast3rslam}, 2025
       & \makecell[c]{6.8/90.14}
       & \makecell[c]{0.35}
       & \fs\bf{0.30}
       & \blackx
       & \blackx
       &
       & \blackx
       & \blackx
       &
       & \makecell[c]{0.31}
       & \makecell[c]{0.40}
       & \blackx
       &
       & \blackx \\

      \bf \ultrafusion (VIO), 2026
      & \nd\sl{3.2/18}
      & \makecell[c]{0.23}
      & \makecell[c]{1.20}
      & \makecell[c]{4.80}
      & \blackx
      &
      & \nd\sl{5.30}
      & \rd{6.50}
      &
      & \nd\sl{0.20}
      & \fs\bf{0.29}
      & \makecell[c]{4.60}
      &
      & \nd\sl{6.90} \\

      \bf \ultrafusion (VWIO), 2026
      & \fs\bf{2.1/2.23}
      & \rd{0.20}
      & \rd{0.66}
      & \rd{1.20}
      & \nd\sl{1.30}
      &
      & \fs\bf{4.50}
      & \makecell[c]{6.70}
      &
      & \fs\bf{0.18}
      & \nd\sl{0.33}
      & \fs\bf{1.05}
      &
      & \fs\bf{6.15}
      \\

       \hline
       A-LOAM\cite{qin2018aloam}, 2018 & \makecell{13.6/13.22} & \makecell[c]{0.15} & \makecell[c]{0.16} & \makecell[c]{6.36} & \makecell[c]{0.19} & & \makecell[c]{66.66} & \makecell[c]{48.37} & & \makecell[c]{0.29} & \makecell[c]{0.26} & \makecell[c]{1.29} & & \makecell[c]{8.46} \\

       LeGO-LOAM\cite{shan2018lego}, 2018 & \makecell{19.8/49.01} & \makecell[c]{7.92} & \blackx & \makecell[c]{13.40} & \makecell[c]{6.28} & & \makecell[c]{19.65} & \blackx & & \makecell[c]{5.89} & \makecell[c]{8.40} & \makecell[c]{32.67} & & \makecell[c]{95.91} \\

       LIO-mapping\cite{ye2019tightly}, 2019 & \makecell{18.3/66.53} & \makecell[c]{2.03} & \makecell[c]{1.95} & \blackx & \makecell[c]{2.46} & & \blackx & \blackx & & \makecell[c]{1.05} & \makecell[c]{1.46} & \makecell[c]{56.3} & & \blackx \\

       LIO-SAM\cite{shan2020lio}, 2020 & \makecell{16.7/38.99} & \makecell[c]{5.10} & \makecell[c]{2.24} & \blackx & \makecell[c]{1.31} & & \makecell[c]{36.15} & \blackx & & \makecell[c]{0.73} & \makecell[c]{0.63} & \makecell[c]{0.56} & & \makecell[c]{43.21} \\

       LINS\cite{qin2020lins}, 2020 & \makecell{20.2/58.05} & \makecell[c]{10.18} & \makecell[c]{3.40} & \makecell[c]{13.25} & \makecell[c]{5.07} & & \blackx & \blackx & & \makecell[c]{5.03} & \makecell[c]{5.04} & \makecell[c]{88.55} & & \blackx \\

       LOAM-Livox\cite{lin2020loam}, 2020 & \makecell{17.9/30.01} & \makecell[c]{2.88} & \makecell[c]{3.24} & \makecell[c]{2.55} & \makecell[c]{2.42} & & \makecell[c]{43.91} & \makecell[c]{87.52} & & \makecell[c]{1.47} & \makecell[c]{1.78} & \makecell[c]{4.36} & & \blackx \\

       LiLi-OM\cite{li2021towards}, 2021 & \makecell{17.3/47.93} & \makecell[c]{1.49} & \makecell[c]{0.19} & \blackx & \makecell[c]{7.00} & & \blackx & \blackx & & \makecell[c]{0.35} & \makecell[c]{2.08} & \makecell[c]{2.22} & & \makecell[c]{15.96} \\

       LIO-Livox\cite{Livox2021LIO}, 2021 & \makecell{14.7/32.04} & \makecell[c]{0.18} & \makecell[c]{0.72} & \makecell[c]{0.30} & \makecell[c]{0.47} & & \blackx & \blackx & & \makecell[c]{0.35} & \makecell[c]{11.25} & \makecell[c]{0.54} & & \makecell[c]{6.63} \\

       Faster-LIO\cite{bai2022faster}, 2022 & \makecell{11.4/31.17} & \rd{0.12} & \makecell[c]{0.13} & \makecell[c]{0.17} & \rd{0.11} & & \blackx & \blackx & & \makecell[c]{2.19} & \makecell[c]{2.84} & \makecell[c]{0.5} & & \makecell[c]{5.60} \\

       IESKF-LIO\cite{chengwei2022iESKFlio}, 2022 & \makecell{6.9/16.59} & \makecell[c]{0.14} & \makecell[c]{0.14} & \rd{0.15} & \makecell[c]{0.13} & & \makecell[c]{14.38} & \blackx & & \makecell[c]{0.17} & \makecell[c]{0.16} & \makecell[c]{0.54} & & \nd\sl{0.08} \\

       VoxelMap\cite{yuan2022efficient}, 2022 & \makecell{10.9/3.07} & \makecell[c]{0.89} & \makecell[c]{0.76} & \makecell[c]{4.93} & \makecell[c]{0.91} & & \makecell[c]{1.08} & \makecell[c]{19.22} & & \makecell[c]{0.99} & \makecell[c]{1.11} & \makecell[c]{0.78} & & \fs\bf{0.07} \\

       FAST-LIO2\cite{xu2022fast}, 2022 & \makecell{9.7/31.49} & \makecell[c]{0.13} & \rd{0.11} & \makecell[c]{0.24} & \rd{0.11} & & \blackx & \blackx & & \rd{0.16} & \makecell[c]{0.18} & \makecell[c]{0.51} & & \makecell[c]{13.46} \\

       CTLO\cite{chengwei2023CTLO}, 2023 & \makecell{6.5/5.79} & \nd\sl{0.10} & \makecell[c]{0.12} & \rd{0.15} & \makecell[c]{0.14} & & \makecell[c]{3.29} & \makecell[c]{52.31} & & \makecell[c]{0.18} & \makecell[c]{0.15} & \makecell[c]{0.54} & & \makecell[c]{0.88} \\

       Point-LIO\cite{he2023point}, 2023 & \makecell{10.8/17.96} & \makecell[c]{0.14} & \makecell[c]{0.14} & \makecell[c]{0.29} & \makecell[c]{0.15} & & \makecell[c]{9.58} & \blackx & & \makecell[c]{0.19} & \makecell[c]{0.20} & \makecell[c]{0.52} & & \makecell[c]{18.39} \\

       LOG-LIO\cite{huang2023log}, 2023 & \makecell{10.6/32.92} & \makecell[c]{0.13} & \makecell[c]{0.12} & \makecell[c]{0.99} & \makecell[c]{0.14} & & \blackx & \blackx & & \makecell[c]{0.18} & \fs\bf{0.07} & \makecell[c]{0.53} & & \makecell[c]{27.03} \\

       CT-LIO\cite{chengwei2023CT}, 2023 & \makecell{6.3/1.00} & \rd{0.12} & \makecell[c]{0.12} & \makecell[c]{0.18} & \makecell[c]{0.13} & & \makecell[c]{3.56} & \makecell[c]{2.39} & & \fs\bf{0.13} & \nd\sl{0.10} & \makecell[c]{0.55} & & \makecell[c]{2.71} \\

       DLIO\cite{chen2023direct}, 2023 & \makecell{8/9.24} & \rd{0.12} & \nd\sl{0.10} & \makecell[c]{0.16} & \makecell[c]{0.15} & & \makecell[c]{40.46} & \makecell[c]{44.70} & & \makecell[c]{0.18} & \makecell[c]{0.17} & \makecell[c]{0.51} & & \makecell[c]{5.80} \\

       HM-LIO\cite{chengwei2023hmlio}, 2023 & \makecell{6.8/15.48} & \rd{0.12} & \makecell[c]{0.15} & \makecell[c]{0.17} & \makecell[c]{0.14} & & \makecell[c]{3.23} & \blackx & & \nd\sl{0.14} & \makecell[c]{0.20} & \makecell[c]{0.53} & & \rd{0.09} \\

       KISS-ICP\cite{vizzo2023ral}, 2023 & \makecell{10.2/30.29} & \makecell[c]{0.15} & \makecell[c]{0.15} & \rd{0.15} & \makecell[c]{0.17} & & \blackx & \blackx & & \makecell[c]{0.25} & \makecell[c]{0.22} & \nd\sl{0.53} & & \makecell[c]{1.3} \\

       SLICT\cite{nguyen2023slict}, 2023 & \makecell{9.6/30.24} & \makecell[c]{0.14} & \makecell[c]{0.14} & \makecell[c]{0.17} & \makecell[c]{0.16} & & \blackx & \blackx & & \makecell[c]{0.25} & \makecell[c]{0.20} & \nd\sl{0.53} & & \makecell[c]{0.81} \\

       MM-LINS\cite{ma2024mm}, 2024 & \makecell{16.5/24.63} & \makecell[c]{2.84} & \makecell[c]{2.79} & \makecell[c]{0.27} & \makecell[c]{1.95} & & \blackx & \makecell[c]{74.31} & & \makecell[c]{2.25} & \makecell[c]{2.83} & \makecell[c]{1.29} & & \makecell[c]{7.73} \\

       SLICT2\cite{nguyen2024eigen}, 2024 & \makecell{8.2/15.44} & \makecell[c]{0.14} & \makecell[c]{0.14} & \makecell[c]{0.17} & \makecell[c]{0.16} & & \makecell[c]{0.87} & \blackx & & \makecell[c]{0.24} & \makecell[c]{0.20} & \nd\sl{0.53} & & \makecell[c]{1.91} \\

       PIN-SLAM\cite{pan2024tro}, 2024 & \makecell{8.4/30.15} & \makecell[c]{0.13} & \makecell[c]{0.14} & \fs\bf{0.05} & \makecell[c]{0.16} & & \blackx & \blackx & & \makecell[c]{0.25} & \makecell[c]{0.18} & \nd\sl{0.53} & & \makecell[c]{0.10} \\

       I2EKF-LO\cite{Yu2024I2EKFLOAD}, 2024 & \makecell{6.9/3.49} & \makecell[c]{2.37} & \makecell[c]{0.13} & \makecell[c]{0.16} & \makecell[c]{0.15} & & \makecell[c]{1.31} & \makecell[c]{29.7} & & \makecell[c]{0.24} & \makecell[c]{0.20} & \nd\sl{0.53} & & \nd\sl{0.08} \\

       LTAOM\cite{zou2024ltaom}, 2024 & \makecell{8/15.20} & \makecell[c]{0.16} & \makecell[c]{0.17} & \makecell[c]{0.18} & \makecell[c]{0.18} & & \rd{0.19} & \blackx & & \makecell[c]{0.26} & \makecell[c]{0.25} & \rd{0.54} & & \nd\sl{0.08} \\

       LOG-LIO2\cite{huang2024loglio2lidarinertialodometryefficient}, 2024 & \makecell{9.7/30.18} & \makecell[c]{0.15} & \makecell[c]{0.15} & \makecell[c]{0.18} & \makecell[c]{0.17} & & \blackx & \blackx & & \makecell[c]{0.25} & \makecell[c]{0.22} & \rd{0.54} & & \rd{0.09} \\

       Eq-LIO\cite{tao2024eqlio}, 2024 & \makecell{6.7/15.19} & \makecell[c]{0.14} & \makecell[c]{0.14} & \makecell[c]{0.18} & \makecell[c]{0.16} & & \makecell[c]{0.20} & \blackx & & \makecell[c]{0.24} & \makecell[c]{0.20} & \nd\sl{0.53} & & \fs\bf{0.07} \\

       Traj-LO\cite{zheng2024traj}, 2024 & \makecell{7.1/15.36} & \makecell[c]{0.13} & \makecell[c]{0.13} & \rd{0.15} & \makecell[c]{0.15} & & \makecell[c]{0.87} & \blackx & & \makecell[c]{0.24} & \makecell[c]{0.19} & \fs\bf{0.52} & & \makecell[c]{1.18} \\

       VoxelMap++\cite{10321658}, 2024 & \makecell{12.3/30.55} & \makecell[c]{0.21} & \makecell[c]{0.18} & \makecell[c]{0.22} & \makecell[c]{0.22} & & \blackx & \blackx & & \makecell[c]{0.24} & \makecell[c]{0.20} & \makecell[c]{1.23} & & \makecell[c]{3.03} \\

       DMSA-SLAM\cite{10610818}, 2024 & \makecell{10.8/45.2} & \makecell[c]{0.13} & \makecell[c]{0.13} & \blackx & \makecell[c]{0.15} & & \blackx & \blackx & & \makecell[c]{0.24} & \makecell[c]{0.20} & \makecell[c]{0.56} & & \makecell[c]{0.58} \\

       Adaptive-LIO\cite{10806842}, 2025 & \nd\sl{3.6/0.23} & \nd\sl{0.10} & \nd\sl{0.10} & \makecell[c]{0.17} & \nd\sl{0.10} & & \makecell[c]{0.26} & \rd{0.57} & & \makecell[c]{0.18} & \rd{0.13} & \rd{0.54} & & \makecell[c]{0.10} \\

       GLO\cite{10891038}, 2025 & \makecell{11.7/31.24} & \makecell[c]{0.15} & \makecell[c]{0.15} & \makecell[c]{0.23} & \makecell[c]{0.16} & & \blackx & \blackx & & \makecell[c]{0.23} & \makecell[c]{0.20} & \makecell[c]{0.63} & & \makecell[c]{10.62} \\

       LIGO\cite{he2025ligo}, 2025 & \makecell{9.1/6.62} & \nd\sl{0.10} & \makecell[c]{0.12} & \makecell[c]{0.25} & \makecell[c]{0.16} & & \makecell[c]{9.55} & \makecell[c]{43.34} & & \makecell[c]{0.19} & \makecell[c]{0.17} & \makecell[c]{0.55} & & \makecell[c]{11.75} \\

       CTE-MLO\cite{shen2025cte}, 2025 & \makecell{8.2/30.16} & \makecell[c]{0.13} & \makecell[c]{0.13} & \rd{0.15} & \makecell[c]{0.15} & & \blackx & \blackx & & \makecell[c]{0.25} & \makecell[c]{0.20} & \nd\sl{0.53} & & \fs\bf{0.07} \\

       RKO-LIO\cite{malladi2025arxiv}, 2025 & \makecell{8.2/15.41} & \makecell[c]{0.15} & \makecell[c]{0.14} & \makecell[c]{0.17} & \makecell[c]{0.16} & & \makecell[c]{2.03} & \blackx & & \makecell[c]{0.25} & \makecell[c]{0.21} & \rd{0.54} & & \makecell[c]{0.48} \\

       II-NVM\cite{10966190}, 2025 & \makecell{10/4.17} & \makecell[c]{0.14} & \makecell[c]{0.14} & \makecell[c]{3.77} & \makecell[c]{0.16} & & \makecell[c]{19.00} & \makecell[c]{14.88} & & \makecell[c]{0.24} & \makecell[c]{0.20} & \makecell[c]{0.56} & & \makecell[c]{2.62} \\

       Surfel-LIO\cite{choi2025surfelliofastlidarinertialodometry}, 2025 & \makecell{9.2/15.51} & \makecell[c]{0.16} & \makecell[c]{0.16} & \makecell[c]{0.18} & \makecell[c]{0.18} & & \makecell[c]{2.87} & \blackx & & \makecell[c]{0.26} & \makecell[c]{0.25} & \makecell[c]{0.55} & & \makecell[c]{0.44} \\

       GenZ-ICP\cite{lee2024genzicp}, 2025 & \makecell{6.4/0.41} & \makecell[c]{0.14} & \makecell[c]{0.14} & \rd{0.15} & \makecell[c]{0.16} & & \makecell[c]{0.32} & \makecell[c]{1.61} & & \makecell[c]{0.25} & \makecell[c]{0.21} & \nd\sl{0.53} & & \makecell[c]{0.63} \\

       Voxel-SLAM\cite{liu2026voxel}, 2026 & \makecell{4.6/0.55} & \fs\bf{0.09} & \fs\bf{0.09} & \makecell[c]{0.18} & \fs\bf{0.09} & & \makecell[c]{1.41} & \makecell[c]{1.45} & & \rd{0.16} & \nd\sl{0.10} & \rd{0.54} & & \makecell[c]{1.39} \\

      \bf \ultrafusion (LIO), 2026 & \rd{3.8/0.19} & \rd{0.12} & \makecell[c]{0.14} & \nd\sl{0.08} & \makecell[c]{0.15} & & \nd\sl{0.17} & \fs\bf{0.25} & & \nd\sl{0.14} & \makecell[c]{0.20} & \rd{0.54} & & \fs\bf{0.07} \\

      \bf \ultrafusion (LWIO), 2026 & \fs\bf{3.5/0.17} & \rd{0.12} & \makecell[c]{0.12} & \nd\sl{0.08} & \makecell[c]{0.14} & & \fs\bf{0.03} & \nd\sl{0.29} & & \nd\sl{0.14} & \makecell[c]{0.20} & \rd{0.54} & & \nd\sl{0.08} \\

       \hline


        LVI-SAM\cite{10.1109/ICRA48506.2021.9561996}, 2021
        & \makecell{6.4/22.88}
        & \makecell[c]{0.85}
        & \makecell[c]{136.03}
        & \makecell[c]{4.23}
        & \makecell[c]{30.85}
        &
        & \makecell[c]{7.06}
        & \makecell[c]{28.44}
        &
        & \makecell[c]{0.64}
        & \rd{0.49}
        & \makecell[c]{7.63}
        &
        & \makecell[c]{12.56}\\

        R2LIVE\cite{lin2021r2live}, 2021
        & \rd{3.2/30.34}
        & \fs\bf{0.11}
        & \fs\bf{0.11}
        & \rd{0.13}
        & \fs\bf{0.10}
        &
        & \blackx
        & \blackx
        &
        & \fs\bf{0.09}
        & \fs\bf{0.19}
        & \makecell[c]{1.33}
        &
        & \makecell[c]{1.36}\\

        R3LIVE\cite{lin2021r3live}, 2022
        & \makecell{6.9/33.8}
        & \makecell[c]{8.76}
        & \makecell[c]{4.24}
        & \makecell[c]{1.12}
        & \makecell[c]{9.00}
        &
        & \makecell[c]{6.07}
        & \blackx
        &
        & \makecell[c]{1.07}
        & \makecell[c]{6.00}
        & \makecell[c]{1.69}
        &
        & \blackx\\

        FAST-LIVO\cite{zheng2022fast}, 2022
        & \makecell{7.3/63.06}
        & \blackx
        & \makecell[c]{8.95}
        & \blackx
        & \makecell[c]{9.49}
        &
        & \makecell[c]{7.96}
        & \blackx
        &
        & \makecell[c]{0.78}
        & \makecell[c]{1.92}
        & \makecell[c]{1.50}
        &
        & \blackx\\

        Coco-LIC\cite{lang2023coco}, 2023
        & \makecell{5.3/16.61}
        & \makecell[c]{1.77}
        & \makecell[c]{0.97}
        & \makecell[c]{0.54}
        & \makecell[c]{1.66}
        &
        & \makecell[c]{6.98}
        & \blackx
        &
        & \makecell[c]{0.64}
        & \makecell[c]{1.80}
        & \rd{1.21}
        &
        & \makecell[c]{0.54}\\

        SR-LIVO\cite{yuan2024sr}, 2024
        & \makecell{6.5/69.59}
        & \makecell[c]{1.23}
        & \makecell[c]{0.28}
        & \nd\sl{0.09}
        & \makecell[c]{1.31}
        &
        & \blackx
        & \blackx
        &
        & \makecell[c]{0.86}
        & \blackx
        & \blackx
        &
        & \makecell[c]{92.14}\\

        FAST-LIVO2\cite{zheng2024fast}, 2024
        & \makecell{4.5/16.57}
        & \makecell[c]{0.44}
        & \makecell[c]{0.28}
        & \makecell[c]{0.17}
        & \makecell[c]{0.33}
        &
        & \rd{3.35}
        & \blackx
        &
        & \makecell[c]{0.51}
        & \makecell[c]{0.81}
        & \makecell[c]{9.71}
        &
        & \rd{0.09}\\

        Ground-Fusion++\cite{zhang2025towards},2025
        & \makecell{3.5/1.1}
        & \rd{0.13}
        & \rd{0.14}
        & \makecell[c]{0.17}
        & \makecell[c]{0.16}
        &
        & \makecell[c]{5.69}
        & \rd{2.48}
        &
        & \rd{0.20}
        & \nd\sl{0.22}
        & \makecell[c]{1.39}
        &
        & \makecell[c]{0.40}\\

        \bf{\ultrafusion (LVIO)}, 2026
        & \nd\sl{1.8/0.22}
        & \nd\sl{0.12}
        & \nd\sl{0.13}
        & \fs\bf{0.08}
        & \rd{0.15}
        &
        & \nd\sl{0.47}
        & \nd\sl{0.28}
        &
        & \nd\sl{0.14}
        & \fs\bf{0.19}
        & \fs\bf{0.53}
        &
        & \nd\sl{0.08}\\

        \bf{\ultrafusion (LVWIO)}, 2026
        & \fs\bf{1.4/0.15}
        & \nd\sl{0.12}
        & \fs\bf{0.11}
        & \fs\bf{0.08}
        & \nd\sl{0.13}
        &
        & \fs\bf{0.02}
        & \fs\bf{0.06}
        &
        & \nd\sl{0.14}
        & \fs\bf{0.19}
        & \nd\sl{0.54}
        &
        & \fs\bf{0.07}\\

        \hline
    \end{tabular}
    \end{adjustbox}
    \noindent\begin{minipage}{1.95\columnwidth}
    \footnotesize{$^1$ In the Avg. Rank/ATE summary, failed runs are counted as 150 m.}
    \end{minipage}
\end{table*}

\begin{table*}[htbp] 
\caption{Comparison of ATE RMSE (m) / drift rate on the M2DGR-Plus dataset~\cite{yin2024ground}.} 
\label{time_asynchronous_tab} 
\renewcommand{\arraystretch}{1.3} 
\begin{adjustbox}{width=2\columnwidth} 
\centering 
\begin{tabular}{*{11}c} 
\hline 
\textbf{Method/Scenario} & {Anomaly} & {Switch} & {Tree} & {Building1} & {Building2} & {Bridge2} & {Parking1} & {Parking2} & {Street1} & {Street2} \\ 
\hline 
FAST-LIVO\cite{zheng2022fast} 
& \blackx & \blackx & \blackx & \blackx & \blackx & \blackx 
&  \makecell[c]{1.06 / 3.80\%} & \blackx 
& \rd \makecell[c]{0.38 / 2.53\%} & \blackx \\ 

FAST-LIVO2\cite{zheng2024fast} 
& \nd\sl \makecell[c]{0.10 / 1.36\%} 
& \nd\sl \makecell[c]{1.60 / 1.86\%} 
& \rd \makecell[c]{2.70 / 3.73\%} 
& \rd \makecell[c]{1.62 / 3.51\%} 
& \blackx 
& \rd \makecell[c]{4.99 / 2.34\%} 
& \rd \makecell[c]{0.81 / 2.90\%} 
& \rd \makecell[c]{0.54 / 1.27\%} 
& \nd\sl \makecell[c]{0.35 / 2.33\%} 
& \rd \makecell[c]{0.57 / 1.56\%} \\ 

Ground-Fusion\cite{yin2024ground}
& \rd \makecell[c]{0.29 / 3.96\%} 
& \rd \makecell[c]{1.80 / 2.09\%} 
& \blackx 
& \nd\sl \makecell[c]{0.60 / 1.29\%} 
& \nd\sl \makecell[c]{0.29 / 1.04\%} 
& \nd\sl \makecell[c]{2.37 / 1.11\%} 
& \nd\sl\makecell[c]{0.48 / 1.73\%}
& \nd\sl \makecell[c]{0.19 / 0.44\%} 
& \makecell[c]{0.47 / 3.13\%}
& \nd\sl \makecell[c]{0.23 / 0.63\%} \\ 

Ground-Fusion++\cite{zhang2025towards}
& \blackx & \blackx 
& \nd\sl \makecell[c]{3.40 / 4.70\%} 
& \blackx 
& \rd \makecell[c]{1.68 / 6.00\%} 
& \blackx 
& \blackx 
& \blackx 
& \blackx 
& \blackx \\ 

\textbf{\ultrafusion (LVWIO)} 
& \fs\bf \makecell[c]{0.09 / 1.23\%} 
& \fs\bf \makecell[c]{0.23 / 0.27\%} 
& \fs\bf \makecell[c]{0.16 / 0.21\%} 
& \fs\bf \makecell[c]{0.32 / 0.69\%} 
& \fs\bf \makecell[c]{0.26 / 0.93\%} 
& \fs\bf \makecell[c]{0.81 / 0.38\%} 
& \fs\bf \makecell[c]{0.10 / 0.36\%} 
& \fs\bf \makecell[c]{0.09 / 0.21\%} 
& \fs\bf \makecell[c]{0.20 / 1.33\%} 
& \fs\bf \makecell[c]{0.12 / 0.33\%} \\
\hline 
\end{tabular} 
\end{adjustbox} 
\vspace{-3mm}
\end{table*}

\subsection{Module Ablations}

\begin{figure}[t]
    \centering
    \includegraphics[width=0.98\columnwidth]{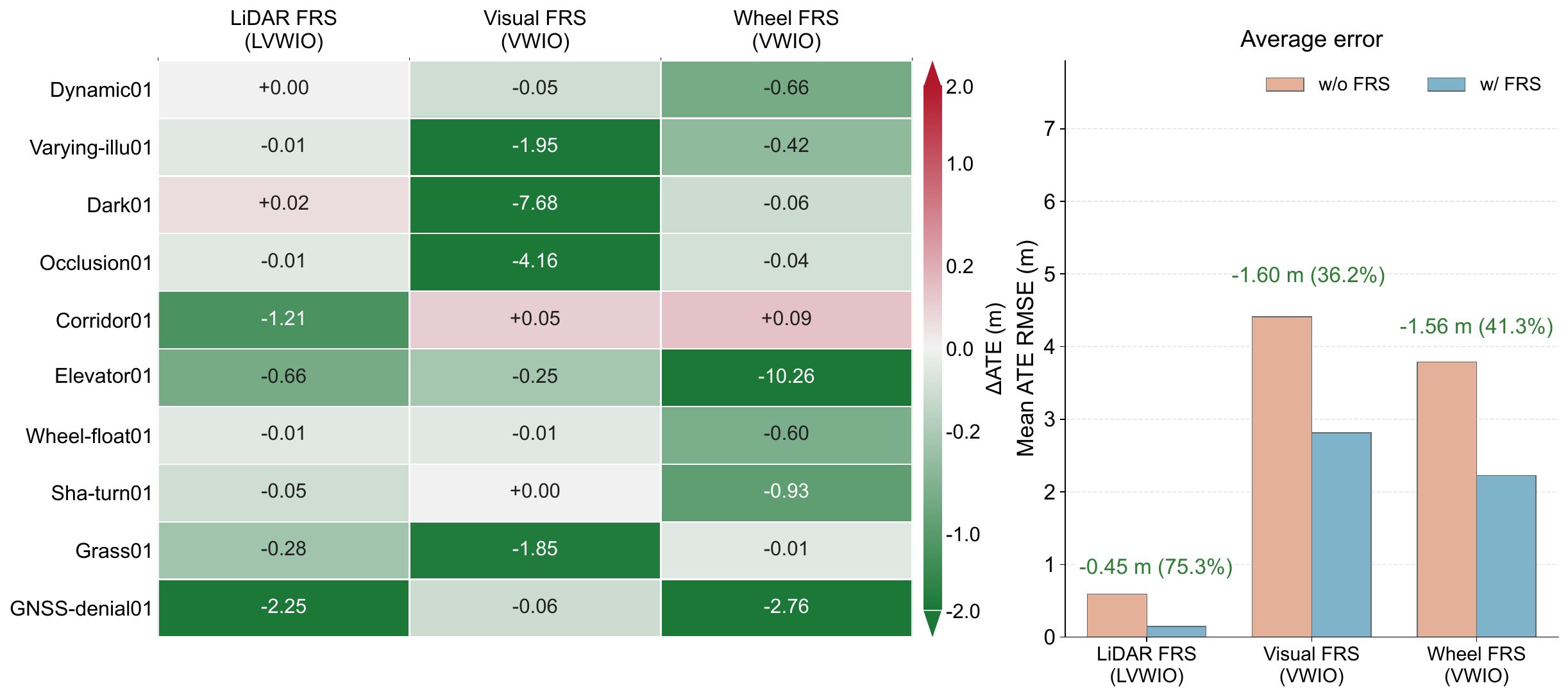}
    \vspace{-3mm}
    \caption{Ablation of Factor-Wise Reliability Scheduling (FRS) on \mthreeDGR. }
    \label{fig:frs_ablation_m3dgr}
    \vspace{-3mm}
\end{figure}

\noindent\textbf{Reliability Scheduling}
Figure~\ref{fig:frs_ablation_m3dgr} compares \ultrafusion with and without each FRS component while keeping the sensor stream active. LiDAR FRS is evaluated in LVWIO, whereas visual and wheel FRS are evaluated in VWIO to avoid masking by dominant LiDAR constraints. Enabling FRS reduces mean ATE by 0.45~m for LiDAR (75.3\%), 1.60~m for vision (36.2\%), and 1.56~m for wheel odometry (41.3\%). GNSS gating yields negligible change on Dark01 (0.0628~m), improves Grass01 from 0.618~m to 0.539~m, and reduces GNSS-denial01 from 2.77~m to 1.79~m, indicating selective rejection of unreliable satellite updates.

\noindent\textbf{Initialization}
Observability-Aware Initialization is ablated by disabling the adaptive bootstrap path and retaining only static IMU/gravity initialization, with pose output suppressed until LiDAR-odometry-aided MAP initialization succeeds (\ultrafusion\ with adaptive initialization). Table~\ref{tab:init_20s_summary} summarizes 18 sequences with complete 20\,s outputs across LVIG, KAIST, \mthreeDGR, M2DGR-Plus, and GrandTour. The full system achieves mean initialization latency of 0.153\,s, median latency of 0.150\,s, and mean 20\,s ATE of 0.483~m, with the fastest initialization in 15/18 sequences and the lowest 20\,s ATE in 11/18. Disabling adaptive initialization raises mean latency to 4.642\,s and mean 20\,s ATE to 16.808~m.

\begin{table}[htbp]
\vspace{-4mm}
\caption{Initialization latency and early-window accuracy on sequences with complete 20\,s outputs from all compared sequences.}
\label{tab:init_20s_summary}
\renewcommand{\arraystretch}{1.12}
\centering
\begin{adjustbox}{width=\columnwidth}
\begin{tabular}{lccc}
\hline
Method & Mean init (s) & Median init (s) & Mean 20\,s ATE (m) \\
\hline
Ground-Fusion++\cite{zhang2025towards} & 2.118 & 1.058 & 85.217 \\
FAST-LIVO2\cite{zheng2024fast} & 0.913 & 0.671 & 28.687 \\
FAST-LIVO\cite{zheng2022fast} & 1.436 & 1.192 & 73.754 \\
\ultrafusion\ (w/o adaptive initialization) & 4.642 & 2.097 & 16.808 \\
\textbf{\ultrafusion} & \bf\fs{0.153} & \bf\fs{0.150} & \bf\fs{0.483} \\
\hline
\multicolumn{4}{l}{\footnotesize{ Failed runs are assigned an initialization time of 20\,s and ATE of 100\,m.}}
\end{tabular}
\end{adjustbox}
\vspace{-3mm}
\end{table}

\begin{figure}[t]
    \centering
    \includegraphics[width=0.9\columnwidth]{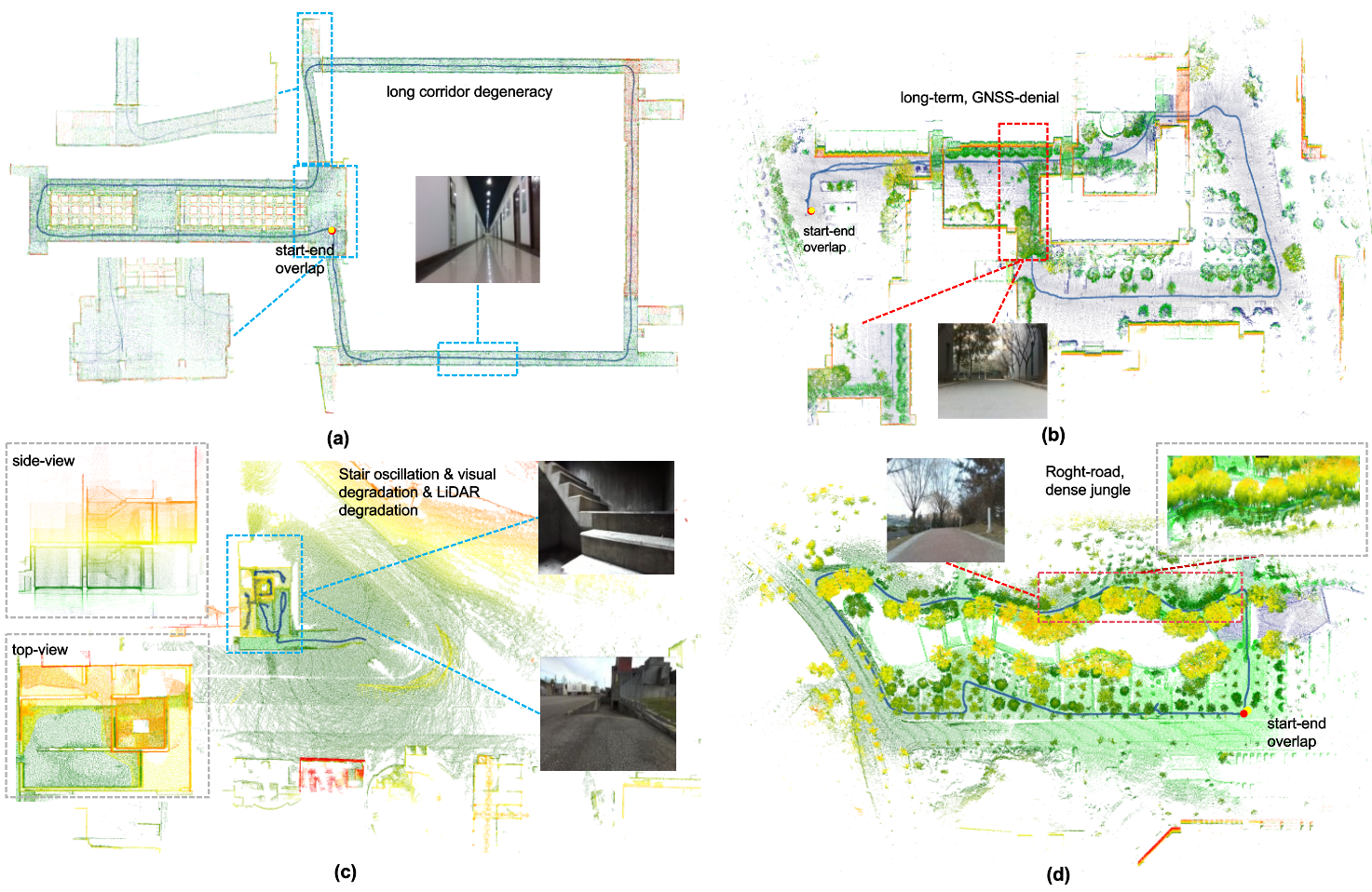}
    \vspace{-2mm}
    \caption{Qualitative robustness under representative sensor-degradation scenarios. Estimated trajectories (blue) and colored point-cloud maps are shown for four stress sequences; red markers denote start--end overlap.
    (a)~\mthreeDGR Corridor01: long-corridor LiDAR degeneracy with weak geometric constraints.
    (b)~\mthreeDGR GNSS-denial01: prolonged GNSS outage with localized visual degradation.
    (c)~GrandTour ARC-2: quadruped stair traversal with body oscillation and concurrent visual/LiDAR degradation.
    (d)~\mthreeDGR Z-Rough-Road01: long-distance navigation in dense vegetation with high visual ambiguity.}
    \label{fig:degradation_qualitative}
    \vspace{-8mm}
\end{figure}

\subsection{Robustness under Sensor Degradation}
The degradation study isolates modality-specific failure mechanisms on \mthreeDGR and complements them with M2DGR-Plus trajectories where multiple degradations may co-occur~\cite{yin2024ground}.

\textbf{Visual degradation:}
Visual degradation weakens feature association and photometric consistency; ORB-SLAM3~\cite{campos2021orb}, TartanVO~\cite{wang2021tartanvo}, and MASt3R-SLAM~\cite{murai2024_mast3rslam} are therefore unstable under darkness or occlusion. \ultrafusion admits visual factors only when feature support, spatial distribution, reprojection consistency, and feature-frame availability are sufficient, allowing LiDAR, inertial, and wheel factors to dominate when visual observability degrades. On long-range routes through dense vegetation, such as Z-Rough-Road01 in Fig.~\ref{fig:degradation_qualitative}(d), heavy occlusion and visual ambiguity can deteriorate correspondence quality; reliability gating suppresses inconsistent visual constraints while complementary modalities preserve trajectory closure.

\textbf{LiDAR degeneracy:}
LiDAR degeneracy stems from rank-deficient geometry. Corridor and elevator scenes reduce constraint diversity, causing FAST-LIO2~\cite{xu2022fast}, FAST-LIVO2~\cite{zheng2024fast}, and SR-LIVO~\cite{yuan2024sr} to degrade or fail. \ultrafusion scores LiDAR factors by geometric conditioning and constraint diversity, enabling LIO/LWIO to down-weight ill-conditioned scans and LVWIO to regularize weak directions with complementary modalities. Fig.~\ref{fig:degradation_qualitative}(a) shows Corridor01, where extended corridor degeneracy weakens observability yet \ultrafusion avoids performance collapse and closes the loop with negligible start--end error.

\begin{figure}[!htbp]
    \centering
    \includegraphics[width=0.9\columnwidth]{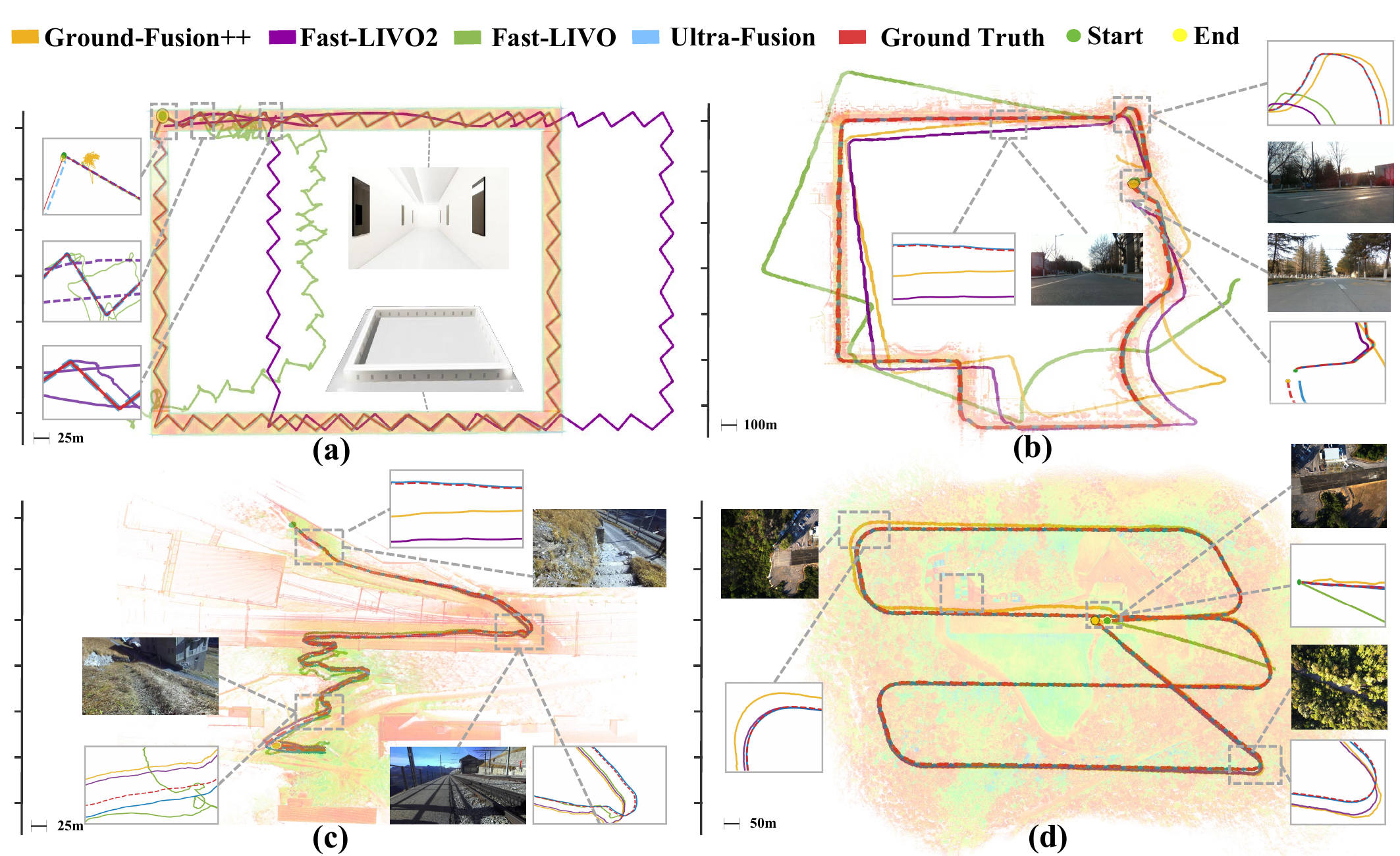}
    \vspace{-2mm}
    \caption{Trajectory estimates on four ITS-relevant sequences: (a) M3DGR Sim Tunnel02, (b) M3DGR Longtime02, (c) GrandTour EIG-1 quadruped, and (d) MARS-LVIG HKAirport02 UAV.}
    \label{fig:trajs}
    \vspace{-4mm}
\end{figure}

The optional intensity cue is evaluated separately. Intensity-consistency residuals are activated only with sufficient local support and serve as a complementary safeguard in geometrically weak regions, rather than replacing geometric registration.
Table~\ref{tab:isaac_lidar_degeneracy} shows that \ultrafusion keeps sub-meter accuracy on Wild scenes and substantially reduces drift in tunnel degeneracy, with the intensity cue improving the Wild sequences and matching geometry-only performance on Tunnel01.
Fig.~\ref{fig:trajs}(a) shows the \mthreeDGR Sim tunnel sequence Tunnel02, where LiDAR-only methods often fail in a geometrically degenerate segment; reliability scheduling down-weights ill-conditioned LiDAR factors while visual and inertial cues preserve continuity.

\begin{table}[!htbp]
    \centering
    \vspace{-2mm}
      \caption{Comparison of ATE RMSE (m) on LiDAR-degenerate Isaac Sim sequences.}
      \label{tab:isaac_lidar_degeneracy}
      \renewcommand{\arraystretch}{1.15}
      \centering
      \scriptsize
      \begin{adjustbox}{width=\columnwidth}
      \begin{tabular}{ccccc}
      \hline
      Method & Wild01 & Wild02 & Tunnel01 & Tunnel02 \\
      \hline
      R3LIVE\cite{lin2021r3live} & 308.03 & 210.97 & \blackx & 1310.17 \\
      FAST-LIVO\cite{zheng2022fast} & \rd 5.31 & 309.44 & \rd 1.05 & 24.64 \\
      FAST-LIVO2\cite{zheng2024fast} & 5.94 & \rd 12.96 & \nd\sl 0.13 & \rd 14.42 \\
      \ultrafusion (w/o intensity) & \nd\sl 0.10 & \nd\sl 0.95 & \fs\bf 0.08 & \fs\bf 2.07 \\
      \bf \ultrafusion (w/ intensity) & \fs\bf 0.06 & \fs\bf 0.70 & \fs\bf 0.08 & \nd\sl 2.21 \\
      \hline
      \end{tabular}
      \end{adjustbox}
\end{table}

\textbf{Wheel slippage:}
Wheel slippage introduces systematic bias that may not be evident from residual magnitude. Raw wheel odometry and Ground-Fusion WIO degrade under slip, weak excitation, or long horizons. \ultrafusion (WIO) reduces average ATE from 35.6~m and 33.68~m to 26.99~m by restricting wheel constraints to their planar observable subspace and validating them with inertial, visual, or LiDAR motion consistency.

\textbf{GNSS denial:}
GNSS denial tests whether global positioning is auxiliary or essential. Methods relying heavily on absolute updates, such as VINS-GPS-Wheel~\cite{Wallong2021} and Ground-Fusion~\cite{yin2024ground}, drift when GNSS becomes unavailable. In \ultrafusion, GNSS is an integrity-checked factor, so local LiDAR, visual, inertial, and wheel constraints continue without reinitialization when GNSS degrades.

\begin{figure}[!hbtp]
    \centering
    \includegraphics[width=0.8\columnwidth]{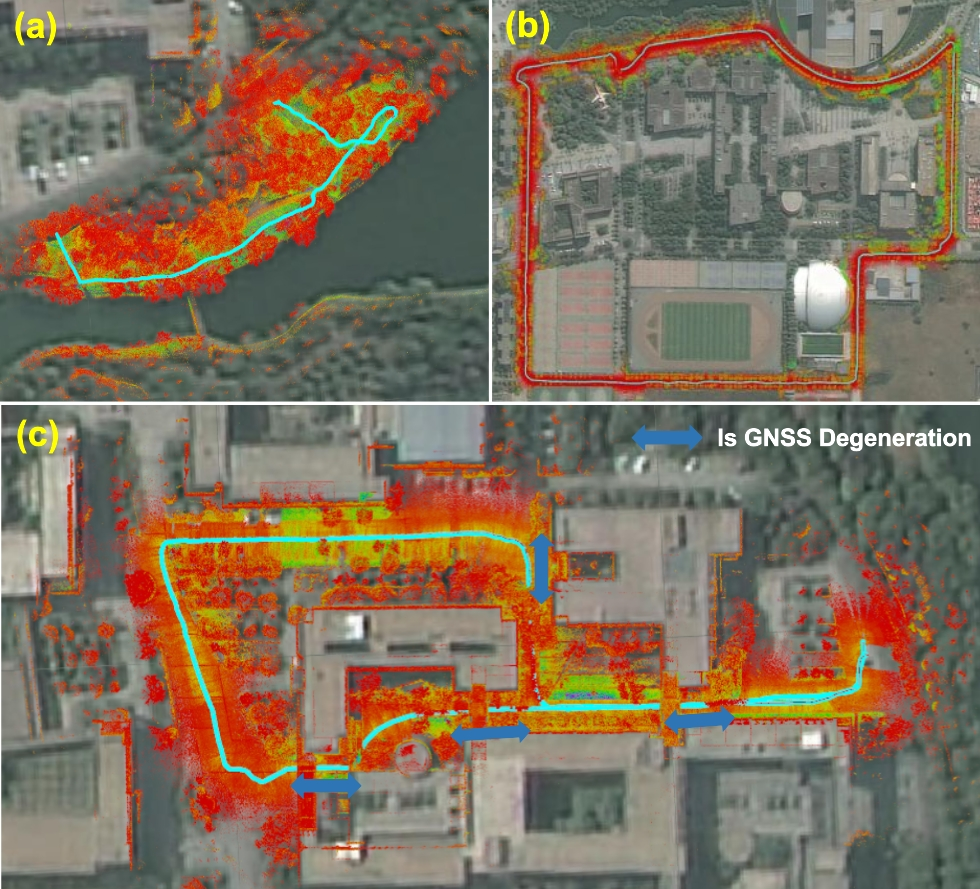}
    \caption{Evaluation of GNSS augmentation and GNSS-denial robustness.
    (a) Grass01 trajectory overlaid on satellite imagery.
    (b) Longtime02 ($>$30~min) trajectory with a large-scale loop.
    (c) GNSS-denial01 sequence, where blue arrows indicate GNSS-denied segments.}
    \label{fig:gnss_qualitative}
    \vspace{-2mm}
\end{figure}

On Longtime02 ($>$30~min), enabling GNSS reduces ATE RMSE from 17.40~m to 8.45~m, indicating that integrity-checked global updates limit drift over extended wheeled patrol trajectories. Fig.~\ref{fig:gnss_qualitative}(b)--(c) shows Longtime02 and GNSS-denial01, where local constraints continue through GNSS-denied segments. Fig.~\ref{fig:degradation_qualitative}(b) further shows accurate convergence to the origin despite prolonged satellite outage and localized visual degradation.

Overall, robustness derives from sensor redundancy and factor-wise reliability scheduling: informative measurements are retained, while inconsistent or degenerate constraints are suppressed. Remaining failures correspond to configurations lacking observations for specific degrees of freedom, rather than forced compensation by low-confidence residuals.

\vspace{-0.5mm}
\subsection{Robustness under Spatiotemporal Miscalibration}

\begin{table}[htbp]
\caption{RMSE (m) under injected IMU time offsets on Wild01.}
\renewcommand{\arraystretch}{1.15}
\centering
\begin{adjustbox}{width=\columnwidth}
\label{tab:preinjected_imu_delay_wild01}
\begin{tabular}{lrrrrrr}
\hline
Time offset & FAST-LIVO\cite{zheng2022fast} & FAST-LIVO2\cite{zheng2024fast} & Ground-Fusion\cite{yin2024ground} & Ground-Fusion{++}\cite{zhang2025towards} & \ultrafusion (w/o OSC$^1$) & \textbf{\ultrafusion (full)} \\
\hline
0ms & 8.6999 & 8.0406 & 3.4194 & 8.8026 & 0.0484 & 0.0319\\
\hline
+100ms & 38.1436 & 10.4990 & 4.1732 & 8.9272 & 0.0718 & 0.0348\\
+200ms & 77.0621 & 9.7028 & 10.0682 & 9.0952 & 0.4684 & 0.0344\\
+300ms & 7.0270 & 6.5562 & 3.5356 & 9.2159 & 0.6919 & 0.0346\\
\hline
-100ms & 118.8689 & 7.8137 & 3.1553 & 9.2141 & 0.0629 & 0.0356\\
-200ms & 157.5297 & 10.8993 & 30.4176 & 8.6085 & 0.1304 & 0.0375 \\
-300ms & 16.7918 & 15.4956 & 3.5999 & 8.4933 & 1.0094 & 0.0403\\
\hline
\multicolumn{7}{l}{\footnotesize{$^1$ Without online LiDAR--IMU temporal calibration (OSC temporal thread).}}
\end{tabular}
\end{adjustbox}
\vspace{-5mm}
\end{table}

\begin{figure}[!t]
    \centering
    \vspace{2mm}
\includegraphics[width=0.9\columnwidth]{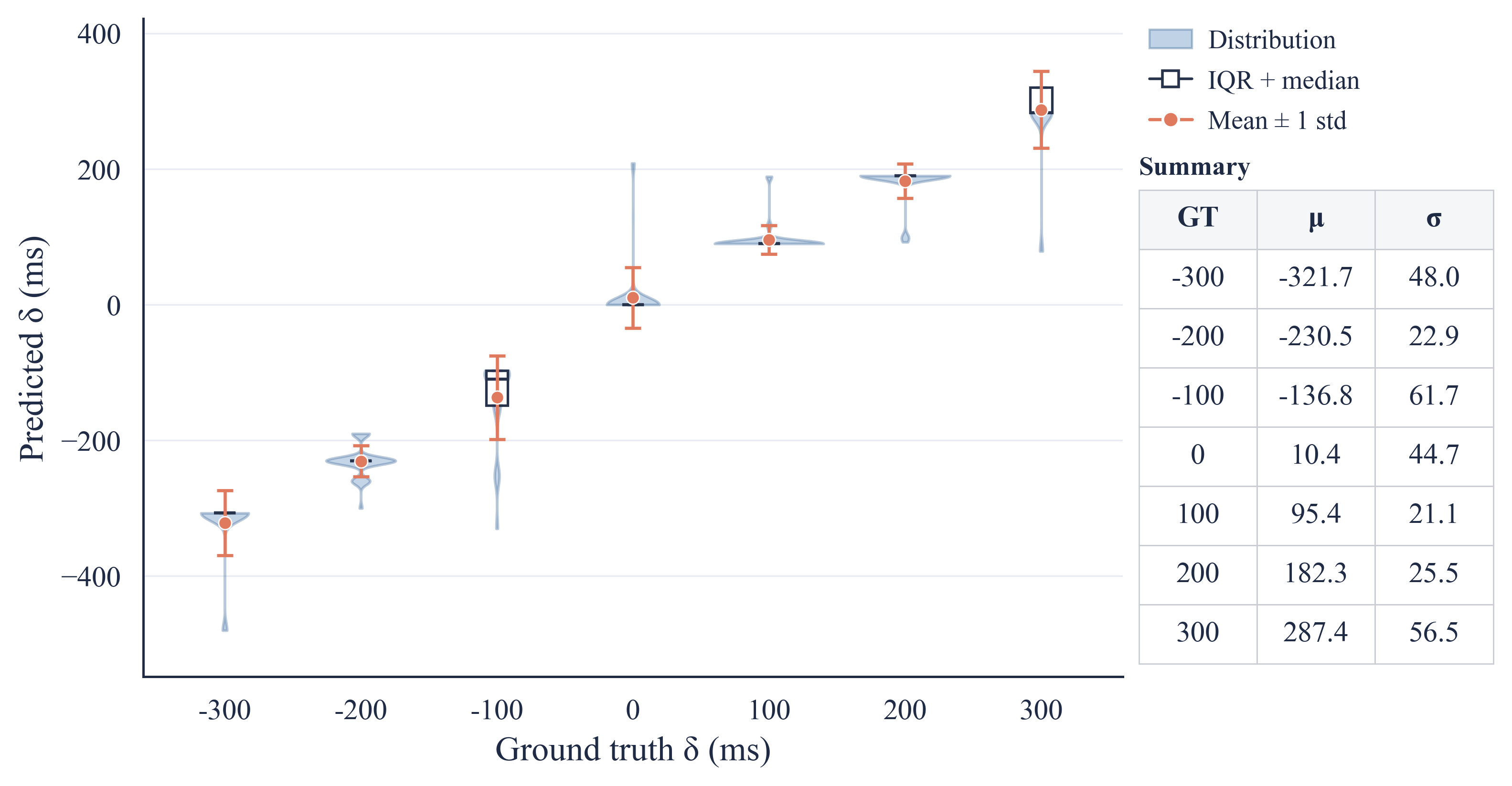}
    \vspace{-2mm}
    \caption{Predicted temporal offsets $\delta$ on Wild01 under injected LiDAR-IMU timing perturbations.}
    \label{fig:imu_delay}
    \vspace{-6mm}
\end{figure}
Spatiotemporal robustness is evaluated by injecting LiDAR--IMU temporal delays and extrinsic rotation perturbations, corresponding to the two parallel OSC threads in Sec.~III-E. Table~\ref{tab:preinjected_imu_delay_wild01} and Fig.~\ref{fig:imu_delay} test the temporal worker on the \mthreeDGR Sim open-area sequence Wild01 with injected IMU delays. \ultrafusion degrades more gradually than the baselines and remains below one meter up to $\pm 200$ ms. Disabling OSC increases error as delay grows, whereas the full system keeps sub-decimeter RMSE and the predicted $\delta$ values concentrate near the injected offsets. Table~\ref{tab:extrinsic_hilti} tests the extrinsic rotation worker on HILTI22\cite{9968057}. FAST-LIVO loses tracking on this fast-motion sequence even at $0^\circ$ perturbation, so its large RMSE reflects trajectory failure rather than rotation sensitivity alone; FAST-LIVO2 and Ground-Fusion++ remain sensitive to injected rotation errors, whereas \ultrafusion maintains sub-meter RMSE over $0$--$10^\circ$ perturbations with online extrinsic calibration enabled.

\begin{table}[htbp]
\centering
\vspace{-3mm}
\caption{RMSE (m) Evaluation on Corridor Lower Gallery 2 from HILTI22 with Extrinsic Rotation Perturbations}
\vspace{-1mm}
\renewcommand{\arraystretch}{1.15}
\begin{adjustbox}{width=\columnwidth}
\label{tab:extrinsic_hilti}
\begin{tabular}{c c c c c c}
\hline
Rotation Error ($^\circ$) & FAST-LIVO\cite{zheng2022fast} & FAST-LIVO2\cite{zheng2024fast} & Ground-Fusion++\cite{zhang2025towards} & \ultrafusion (w/o extr. calib.$^1$) & \textbf{\ultrafusion (full)} \\
\hline
0.00 & \blackx & 0.15 & 2.93 & 0.12 & \bf\fs{0.10}\\
1.00 & \blackx & 0.18 & 3.91 & \bf\fs0.09 & {0.12}\\
2.00 & \blackx & 0.18 & 3.94 & 0.19 & \bf\fs{0.10}\\
3.00 & \blackx & 0.13 & 2.44 & 0.34 & \bf\fs{0.10}\\
4.00 & \blackx & 0.26 & 5.52 & 0.23 & \bf\fs{0.18}\\
5.00 & \blackx & 0.61 & 3.83 & 0.41 & \bf\fs{0.11}\\
6.00 & \blackx & 0.81 & 3.38 & 0.47 & \bf\fs{0.12}\\
7.00 & \blackx & 8.81 & 1.50 & 0.43 & \bf\fs{0.14}\\
8.00 & \blackx & 145.11 & 2.75 & 0.44 & \bf\fs{0.10}\\
9.00 & \blackx & 27.80 & 2.07 & 0.68 & \bf\fs{0.12}\\
10.00 & \blackx & 940.37 & 3.81 & 0.75 & \bf\fs{0.25}\\
\hline
\multicolumn{6}{l}{\footnotesize{$^1$ Without online LiDAR--IMU extrinsic calibration. FAST-LIVO loses tracking on HILTI22 at all perturbation levels.}}
\end{tabular}
\end{adjustbox}
\vspace{-5mm}
\end{table}


\vspace{-0.5mm}
\subsection{Long-Term and High-Speed Operation}

Long-duration \mthreeDGR trajectories and city-scale high-speed KAIST urban-driving sequences evaluate prolonged operation, rapid motion, and accumulated calibration error.
On Longtime01 and Longtime02, both exceeding 30 minutes, \ultrafusion obtains the lowest positioning errors (Table~\ref{tab:long_term_test}), which limits drift accumulation during extended shuttle or patrol missions.

\begin{table}[!hbp]
\centering
\vspace{-4mm}
\caption{Comparison of ATE RMSE (m) on long-duration \mthreeDGR sequences.}
\vspace{-1mm}
\label{tab:long_term_test}
\begin{adjustbox}{width=0.8\columnwidth}

\begin{tabular}{ccc}
\hline
{Methods} & {Longtime01} & {Longtime02}  \\ \hline
{FAST-LIVO\cite{zheng2022fast}}         & 20.5   & 27.5   \\
{FAST-LIVO2\cite{zheng2024fast}}        & \nd \sl 5.13   & \nd \sl 8.4    \\
{Ground-Fusion\cite{yin2024ground}}     & 22.5   & \blackx      \\
{Ground-Fusion++\cite{zhang2025towards}}   & \rd 7.5    &\rd 15.9   \\
{\textbf{Ultra-Fusion (LVWIO)}}& \fs\bf 4.3    &  \fs\bf 2.8    \\
\hline
\end{tabular}

\end{adjustbox}
\vspace{-2mm}
\end{table}

On the city-scale KAIST dataset, with speeds from 25.2 to 96.9 km/h, \ultrafusion (LVWIO) reduces average drift to 0.38\% (Table~\ref{kaist_tab}). FAST-LIVO and FAST-LIVO2 exhibit large drift on these high-speed trajectories. Wheel--inertial motion support and LiDAR/visual drift correction remain complementary when optimized in a common window.
Fig.~\ref{fig:trajs}(b) shows the Longtime02 trajectory over a 30+ minute route with repeated revisits to mapped regions.

\begin{table*}[!hb]
\caption{ATE RMSE (m) / drift rate Comparison on KAIST~\cite{jeong2019complex} dataset. Sequence headers list trajectory length and avg. speed.}
\label{kaist_tab}
\renewcommand{\arraystretch}{1.3}
\centering
\begin{adjustbox}{width=1.8\columnwidth}
\begin{tabular}{cccccc}
\hline
\multirow{2}{*}{Method / Seq. Info.} 
& Urban23 & Urban25 & Urban26 & Urban29 & Urban35 \\
& \footnotesize{3379.7 m / 96.9 km/h} 
& \footnotesize{2505.1 m / 91.4 km/h} 
& \footnotesize{3987.8 m / 25.2 km/h} 
& \footnotesize{3559.0 m / 29.1 km/h} 
& \footnotesize{3187.7 m / 67.0 km/h} \\
\hline

FAST-LIVO\cite{zheng2022fast} 
& 979.84 / 28.99\% 
&  714.29 / 28.51\% 
& 487.91 / 12.24\% 
& 822.62 / 23.11\% 
& 873.49 / 27.40\% \\

FAST-LIVO2 \cite{zheng2024fast}
& 979.40 / 28.98\% 
& 705.77 / 28.17\% 
&  497.03 / 12.46\% 
& 818.37 / 22.99\% 
& 918.62 / 28.82\% \\

Ground-Fusion\cite{yin2024ground} 
& 1825.70 / 54.02\% 
& 486.74 / 19.43\% 
& 173.52 / 4.35\% 
& \blackx 
& 703.14 / 22.06\% \\

Ground-Fusion++\cite{zhang2025towards} 
& 586.28 / 17.35\% 
& \blackx 
& \blackx 
& \blackx 
& \blackx \\

VINS-GPS-Wheel\cite{Wallong2021} 
& \blackx
& \blackx
&  \blackx
&  \blackx
&  \blackx \\

Raw Wheel Odom
& \rd 284.60 / 8.42\%
& \nd\sl 14.63 / 0.58\%
& \nd\sl 36.15 / 0.91\%
& \nd\sl 19.42 / 0.55\%
& \rd 4.44 / 0.14\% \\

\bf \ultrafusion (WIO) 
& \nd\sl 279.90 / 8.28\%
& \rd 14.69 / 0.59\%
& \rd 36.56 / 0.92\%
& \rd 19.88 / 0.56\%
& \nd\sl 4.43 / 0.14\% \\

\bf \ultrafusion (LVIO) 
&  334.77 / 9.91\%
& 441.68 / 17.63\%
& 428 / 10.22\% 
& 525.25 / 14.76\%
& 671.15 / 21.05\% \\

\bf \ultrafusion (LVWIO) 
& \fs\bf 12.38 / 0.37\% 
& \fs\bf 14.56 / 0.37\% 
& \fs\bf 32.50 / 0.58\% 
& \fs\bf 16.43 / 0.46\% 
& \fs\bf 4.12 / 0.13\% \\
\hline
\end{tabular}
\end{adjustbox}
\vspace{-3mm}
\end{table*}

\begin{table*}[!hb]
\caption{Comparison of RTE (cm) on GrandTour~\cite{frey_tuna2026grandtour}$^1$ across representative legged-robot sequences.}
\label{grandtour}
\renewcommand{\arraystretch}{1.15}
\centering
\scriptsize
\begin{adjustbox}{width=1.6\columnwidth}
\begin{tabular}{c|cccc}
\hline
\multirow{2}{*}{Method / Seq. Info.} 
& SPX-2 
& SNOW-2 
& EIG-1 
& ARC-2 \\
& \footnotesize{urban / large-scale} 
& \footnotesize{snowy / low-visibility} 
& \footnotesize{industrial / cluttered} 
& \footnotesize{debris / unstructured} \\
\hline
Traj-LO\cite{zheng2024traj}   &  1.11 & 1.28 &  1.12 & 17.36 \\
DLO\cite{9681177}       & 1.77 & 3.07 & 2.49 & 2.91 \\
I2EKF-LO\cite{Yu2024I2EKFLOAD}  & 1.22 & 4.04 & 4.29 & 3.69 \\
CT-LO\cite{dellenbach2022ct}     & 1.55 & 1.45 & 1.53 & 4.23 \\
CTE-MLO\cite{shen2025cte}   & 2.12 & 1.21 & 1.65 & 11.13 \\
GenZ-ICP\cite{lee2024genzicp}  & 1.51 & 26.40 & 3.75 & 3.09 \\
RESPLE-LO\cite{cao2025resple} & 1.40 & 2.01 & 2.02 & 3.71 \\
KISS-ICP\cite{vizzo2023ral}  & 3.28 & 35.78 & 7.76 & 13.71 \\

Coco-LIC\cite{lang2023coco}     & \nd\sl 0.44 & \nd\sl 0.41 & \nd\sl 0.40 & \rd 1.01 \\
FAST-LIVO\cite{zheng2022fast}   & 69.98 & 69.62 & 75.13 & 63.01 \\
FAST-LIVO2\cite{zheng2024fast}   &  1.05 &  1.25 &  1.11 & \fs\bf 0.70 \\
Ground-Fusion\cite{yin2024ground}   & 983.19 & 6.42 & \blackx & 30.07 \\
Ground-Fusion++\cite{zhang2025towards}   & 1.82 & 1.90 & 3.96 & 2.83 \\
PV-LIO\cite{PVLIORepo}      & 1.14 & 1.32 & 1.24 & 1.05 \\
DLIO\cite{chen2023direct}         & 1.22 & 1.51 & 1.26 & 1.19 \\
Fast-LIMO\cite{FastLIMORepo}    & 1.15 & \rd 1.16 & \rd 1.10 & 2.49 \\
Voxel-SLAM\cite{liu2026voxel}   &\rd 1.03 & 1.37 & 1.34 & 1.24 \\
\bf \ultrafusion (LVIO) & \fs\bf 0.41 & \fs\bf 0.34 & \fs\bf 0.26 & \nd\sl 0.90 \\
\hline
\multicolumn{5}{l}{\footnotesize$^1$  Baseline results are taken from \cite{frey_tuna2026grandtour}.}
\end{tabular}
\end{adjustbox}
\vspace{-5mm}
\end{table*}

\vspace{-0.5mm}
\subsection{Cross-Platform Validation}
\vspace{-0.5mm}
We evaluate whether the Ultra-Fusion framework transfers beyond the wheeled settings. We use GrandTour for legged robots and MARS-LVIG for aerial robots, covering body oscillation, shocks, large viewpoint changes, rapid motion, and weak structural priors.

\noindent\textbf{Legged Robots:}
GrandTour~\cite{frey_tuna2026grandtour} evaluates transfer to quadruped platforms relevant to last-mile delivery and facility inspection, using Relative Trajectory Error over 0.5-meter path segments. Body oscillation, attitude variation, shocks, and rapid viewpoint changes challenge registration and tracking. Table~\ref{grandtour} shows that \ultrafusion obtains the lowest error on three of four sequences and remains competitive on the remaining one, so the estimator is not restricted to wheel-specific motion priors.
Fig.~\ref{fig:trajs}(c) shows the GrandTour EIG-1 quadruped sequence on uneven terrain, where close alignment with ground truth is maintained despite slip-prone contacts and visual occlusion. Fig.~\ref{fig:degradation_qualitative}(c) further reports ARC-2, where up--down stair motion induces platform oscillation and abrupt viewpoint changes that limit visual observability; \ultrafusion remains stable without accumulated drift.

\noindent\textbf{Aerial Robots:}
For aerial validation, MARS-LVIG~\cite{li2024mars} provides UAV trajectories at 80--130 m altitude for corridor inspection, airport surroundings, and low-altitude urban mapping. Large viewpoint variation, rapid motion, and weak structural priors cause several baselines to degrade or fail. Table~\ref{lvig} shows that \ultrafusion achieves low average error, supporting transfer beyond ground-robot motion statistics and yielding georeferenced trajectories suitable for infrastructure monitoring along transportation corridors.
Fig.~\ref{fig:trajs}(d) shows the MARS-LVIG HKAirport02 UAV sequence at about 80~m altitude in an urban airport environment.

\begin{figure}[!htbp]
    \centering
    \vspace{-3mm}
    \includegraphics[width=\columnwidth]{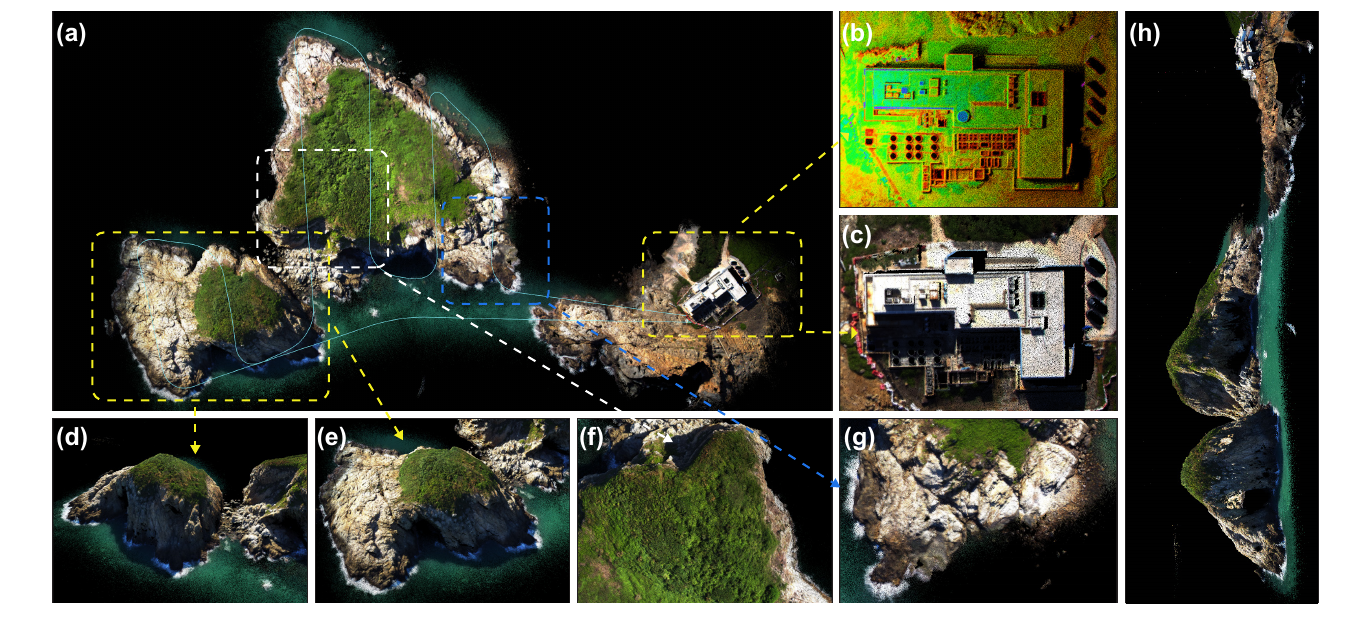}
    \vspace{-7mm}
    \caption{Reconstruction map of the MARS-LVIG HKisland03 sequence.
    (a) Global top-view map with the estimated trajectory and highlighted regions.
    (b)--(c) Intensity-colored and RGB-colored details of the building area.
    (d)--(e) Enlarged views of the lower-left island.
    (f)--(g) Enlarged views of the central island.
    (h) Side-view reconstruction showing the 3D scene structure.}
    \label{fig:lvig_map}
    \vspace{-2mm}
\end{figure}

\begin{table*}[!htbp]
\caption{ATE (m) Comparison on representative MARS-LVIG~\cite{li2024mars} sequences$^1$. 
Sequence headers list flight speed and altitude.}
  \label{lvig}
  \renewcommand{\arraystretch}{1.15}
  \centering
  \scriptsize
  \begin{adjustbox}{width=1.8\columnwidth}
  \begin{tabular}{cccccccc}
  \hline
\multirow{2}{*}{Method / Seq. Info.} 
& \multirow{2}{*}{\makecell{Avg. Rank\\ / RMSE}}
& HKairport01 
& HKairport02 
& HKisland03 
& AMtown03 
& AMvalley03 
& HKGNSS02 \\
&
& \footnotesize{ 3 m/s / 80 m}
& \footnotesize{ 6 m/s / 80 m}
& \footnotesize{ 9 m/s / 90 m}
& \footnotesize{ 12 m/s / 80 m}
& \footnotesize{ 12 m/s / 130 m}
& \footnotesize{ 6 m/s / 80 m} \\
  \hline
  \hline
  LIO-Livox\cite{Livox2021LIO}     & \makecell[c]{6.0/96.17} & \rd 0.65 & 123.39 & \blackx    & \blackx    & \blackx    & 2.97 \\
  FAST-LIO2\cite{xu2022fast}       & \makecell[c]{4.3/2.98}  & \nd\sl 0.44 & 0.96   & 2.13 & 3.65 & 7.77 & 2.92 \\
  iG-LIO\cite{chen2024ig}          & \makecell[c]{6.8/125.59} & \blackx    & \blackx      & \blackx    & 3.54 & \blackx    & \blackx    \\
  R3LIVE\cite{lin2021r3live}       & \makecell[c]{5.3/51.40} & 0.68 & \nd\sl 0.82 & 3.93 & \blackx    & \blackx    & 2.98 \\
  AKF-LIO\cite{xie2025akf}         & \rd\makecell[c]{3.0/2.15} & \fs\bf 0.43 & \rd 0.87 & 2.03 & \fs\bf 2.74 & 3.91 & 2.94 \\
  FAST-LIVO\cite{zheng2022fast}    & \makecell[c]{7.3/150.00} & \blackx & \blackx & \blackx & \blackx & \blackx & \blackx \\
  FAST-LIVO2\cite{zheng2024fast}   & \nd\sl\makecell[c]{2.7/1.42} & 0.75 & 0.88 & \nd\sl 0.89 & \rd 3.25 & \fs\bf 1.43 & \nd\sl 1.32 \\
  Ground-Fusion\cite{yin2024ground} & \makecell[c]{7.3/150.00} & \blackx & \blackx & \blackx & \blackx & \blackx & \blackx \\
  Ground-Fusion++\cite{zhang2025towards} & \makecell[c]{4.5/2.22} & 1.10 & 1.28 & \rd 1.71 & 3.82 & \rd 3.33 & \rd 2.05 \\
  \bf \ultrafusion (LVIO)          & \fs\bf\makecell[c]{1.3/1.40} & \fs\bf 0.43 & \fs\bf 0.61 & \fs\bf 0.87 & \nd\sl 3.01 & \nd\sl 2.60 & \fs\bf 0.90 \\
  \hline
  \multicolumn{8}{l}{\footnotesize$^1$ Baseline results are taken from \cite{li2024mars,xie2025akf, tang2026palviorealtimelidarvisualinertialodometry}.}
  \end{tabular}
  \end{adjustbox}
  \vspace{-7mm}
  \end{table*}

\subsection{Mapping Results}

Map consistency is evaluated qualitatively using the hybrid local map. Fig.~\ref{fig:lvig_map} shows the MARS-LVIG HKisland03 reconstruction, where island contours, road boundaries, buildings, and terrain transitions remain coherent over a large-scale trajectory. The enlarged views show clear structural edges and limited ghosting in both intensity- and RGB-colored point clouds, supporting the use of optimized poses and aligned colorized LiDAR observations for Gaussian mapping.


\subsection{Runtime Analysis}
\begin{figure}[!ht]
    \centering
    \vspace{-4mm}
    \includegraphics[width=\columnwidth]{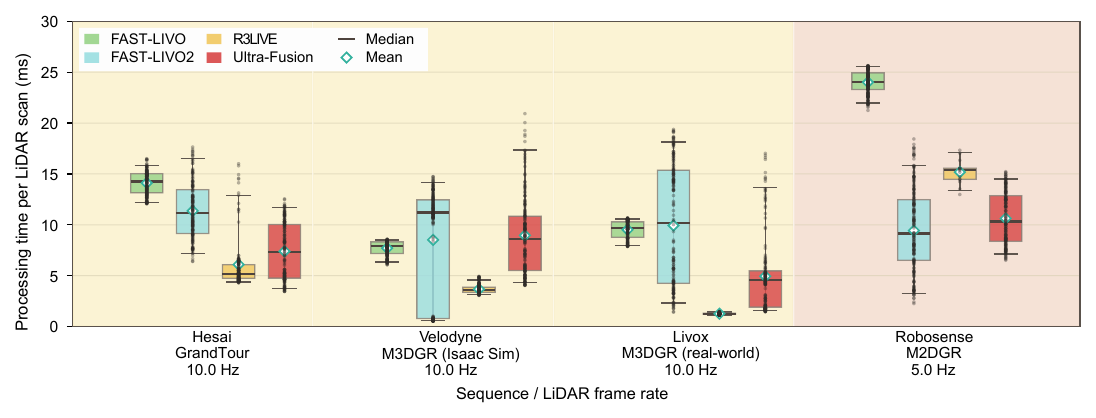}\\
    \vspace{-4mm}
    \caption{Time consumption per LiDAR scan of Ultra-Fusion with baselines on different LiDAR types.}
    \label{fig:time_consumption}
    \vspace{-3mm}
\end{figure}

Runtime is profiled on 60~s segments with Robosense, Velodyne, Livox, and Hesai LiDARs using a real-time configuration: per-frame LiDAR frontend, window size 4, at most 3 nonlinear iterations, 12~ms solver budget, and capped factors on an Intel Core i9-14900K CPU. \ultrafusion requires 5.48--10.73~ms per optimization step (Fig.~\ref{fig:time_consumption}), satisfying real-time operation in this setting.

\section{Conclusion and Limitations}

This paper addresses multi-sensor localization for intelligent transportation systems under sensor degradation and spatiotemporal uncertainty. \ultrafusionc formulates heterogeneous LiDAR, visual, inertial, wheel, and GNSS measurements as factors in a Unified Sliding-Window Estimator, so WIO, VIO, LIO, LVIO, and augmented configurations share state, calibration, reliability scheduling, and marginalization logic. Observability-Aware Initialization, Factor-Wise Reliability Scheduling, and Online Spatiotemporal Calibration improve initialization reliability, degradation tolerance, and calibration robustness.
Extensive evaluation on \mthreeDGRc and public benchmarks shows competitive performance across wheeled, quadruped, and UAV platforms under long-term/high-speed motion and controlled perturbations. 

\noindent\textbf{Limitations:} the study focuses on localization and geometric mapping; semantic scene understanding and explicit dynamic-object reasoning remain outside the evaluated scope. Extending the reliability-aware formulation toward semantic and dynamic environments is an important direction for future transportation deployments.

\section*{Declarations and Data Availability}
\noindent\textbf{Data Availability.}
Most \mthreeDGR sequences are publicly available at \url{https://github.com/sjtuyinjie/M3DGR}. The two long-duration sequences and M3DGR Sim sequences will be released upon acceptance. Executable binaries of \ultrafusion are available at \url{https://github.com/sjtuyinjie/Ultra-Fusion}; source code will be released upon acceptance.

\noindent\textbf{Acknowledgements and Conflict of Interest.}
This work was self-funded. The authors declare that they have no known competing financial interests or personal relationships that could have appeared to influence this work.

\noindent\textbf{Author Contributions.}
{Yihong Tian:} Software, Methodology, Validation.
{Junjie Zhang:} Data curation, Methodology, Software.
{Liuyang Li:} Software, Validation, Visualization.
{Deteng Zhang:} Data curation.
{Yunfei Zuo:} Simulation.
{Jie Yin:} Conceptualization, Methodology, Writing, and Supervision.


\bibliographystyle{IEEEtrans}
\bibliography{root}

\clearpage
\onecolumn
\providecommand{\Checkmark}{{\ding{51}}}

\section*{Supplementary Materials}
In this material, we provide supplementary tables for system configuration, benchmark comparison, and evaluated-framework licensing information.

\subsection{System Illustration}

\begin{figure}[!htbp]
    \centering
    \vspace{-1mm}
    \begin{adjustbox}{width=0.8\linewidth}
    \includegraphics{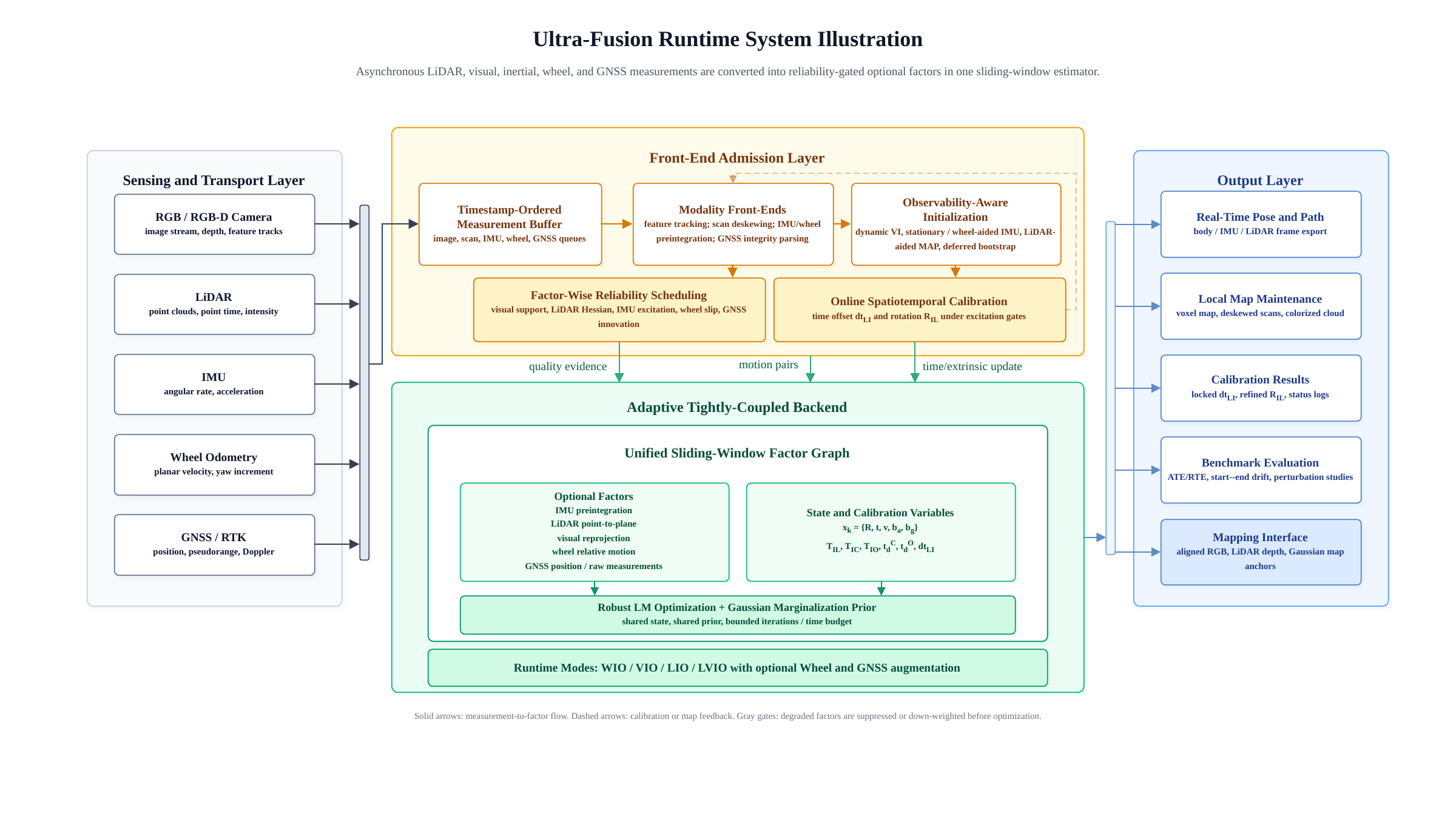}
    \end{adjustbox}
    \vspace{-1mm}
    \caption{System illustration of \ultrafusion. Heterogeneous measurements are timestamp-ordered, screened by modality front-ends, initialized under observability-aware bootstrap, and admitted into a unified sliding-window factor graph through reliability-gated optional factors. The same backend state supports pose estimation, local map maintenance, online LiDAR--IMU spatiotemporal calibration, and benchmark outputs across WIO, VIO, LIO, and LVIO configurations.}
    \label{fig:supp_system_illustration}
\end{figure}

Fig.~\ref{fig:supp_system_illustration} gives a runtime-level illustration of \ultrafusion. The system first aligns asynchronous ROS topics in a timestamp-ordered buffer, then converts the admitted measurements into modality-specific front-end products. Observability-Aware Initialization (OAI), Factor-Wise Reliability Scheduling (FRS), and Online Spatiotemporal Calibration (OSC) operate before or alongside the common backend, so WIO, VIO, LIO, LVIO, and optional wheel/GNSS-augmented modes share the same state, marginalization prior, and solver interface.

\subsection{Key Parameters}

Table~\ref{tab:supp_state_factor_config} maps the runtime blocks in Fig.~\ref{fig:supp_system_illustration} to the main configuration parameters used by \ultrafusion. The listed values follow the default \mthreeDGR configuration; sequence-specific files may adjust noise levels, residual caps, and platform-dependent gates for different LiDAR types or motion regimes.

\begin{table}[!htbp]
\scriptsize
\centering
\caption{Key runtime parameters of \ultrafusion.}
\label{tab:supp_state_factor_config}
\renewcommand{\arraystretch}{1.08}
\setlength{\tabcolsep}{3pt}
\begin{adjustbox}{width=0.9\linewidth}
\begin{tabular}{@{}>{\raggedright\arraybackslash}p{0.15\linewidth}
                >{\raggedright\arraybackslash}p{0.22\linewidth}
                >{\raggedright\arraybackslash}p{0.25\linewidth}
                >{\raggedright\arraybackslash}p{0.31\linewidth}@{}}
\hline
\textbf{Group / block} & \textbf{Parameters} & \textbf{Runtime role} & \textbf{Default values} \\
\hline
\multicolumn{4}{@{}l}{\textbf{Input and front-end processing}}\\
Sensor buffer & Topic association, packet ordering, point filtering & Provides timestamp-ordered packets for all enabled modalities & IMU 100~Hz; image 10~Hz; wheel 10~Hz; LiDAR blind range 0.1~m \\
Visual front-end & Feature count, spacing, track length, parallax & Produces visual tracks and reprojection factors when visual support is sufficient & Max 150 features; min distance 30 px; min track length 4; keyframe parallax 10 px \\
LiDAR front-end / map & Residual cap, voxel size, plane gate, robust loss & Produces scan-to-map factors and LiDAR reliability evidence & Residual cap 1200--2000; voxel 0.5~m; Huber $\sigma=0.5$--1.0; plane eigen-ratio gate 0.12 \\
\hline
\multicolumn{4}{@{}l}{\textbf{Motion support and global anchoring}}\\
IMU / wheel support & IMU noise, wheel noise, stationary checks & Supplies preintegration and slip/stationary evidence & $\sigma_a=0.04$, $\sigma_g=0.008$; wheel velocity noise 0.2--0.6; gravity alignment 1.0~s \\
GNSS anchoring & Satellite/elevation checks, covariance, innovation gate & Admits optional global factor after integrity checks & Elevation $30^\circ$; min satellites 5; innovation gate 5.0; horizontal/vertical noise 5/20~m \\
\hline
\multicolumn{4}{@{}l}{\textbf{Robustness and calibration}}\\
OAI bootstrap & Initial LiDAR frames, IMU window, wheel-aided checks & Selects visual--inertial, stationary/wheel-aided, LiDAR-aided MAP, or deferred initialization & LiDAR init frames 20; IMU gravity init 1.0~s; wheel stop threshold 0.01 \\
FRS gates & LiDAR geometry, visual support, IMU excitation, wheel slip, GNSS innovation & Activates, suppresses, or covariance-inflates factors before backend optimization & Visual min feature 100; LiDAR normal radius 0.5~m; sigma gate $k=3.0$ \\
OSC worker & Time-offset search, overlap duration, confidence, extrinsic lock & Updates LiDAR--IMU time association and rotation under sufficient excitation & $\Delta t_{LI}$ range $\pm0.5$~s; 12~s window; min overlap 5~s; confidence 0.6 \\
\hline
\multicolumn{4}{@{}l}{\textbf{Optimization backend}}\\
Unified backend & Window size, state variables, optional factors, solver budget & Jointly optimizes active WIO/VIO/LIO/LVIO factors with shared marginalization & 10-state window; Ceres LM; 8 iterations; 40~ms budget \\
\hline
\end{tabular}
\end{adjustbox}
\end{table}

\subsection{M3DGR Dataset}

\begin{table}[!htbp]
    \footnotesize
    \centering
    \caption{Comparison of representative SLAM benchmark datasets in terms of scenarios, sensors, and evaluation breadth.}
    \label{Comparison}
    \renewcommand{\arraystretch}{1.12} 
    \begin{adjustbox}{width=0.9\linewidth}
        \begin{tabular}{*{14}c}
            \hline
            \multirow{2}{*}{\makecell{Dataset/Year}} & \multicolumn{5}{c}{\makecell{Scenario}}  & \multicolumn{7}{c}{\makecell{Sensors}} & \multirow{2}{*}{\makecell{Number of \\ compared algorithms }} \\
                \cline{2-5} \cline{7-13}
                & VC$^1$ & LR$^2$ & GD$^3$ & WS$^4$ &  & RGB & Depth & Omni$^5$ & IMU & LiDAR & Wheel & GNSS &  \\
            \hline

            \makecell{EuRoC, 2016} & \makecell{\Checkmark} & \makecell{} & \makecell{} & \makecell{} & \makecell{} & \makecell{\Checkmark} & \makecell{} & \makecell{} & \makecell{\Checkmark} & \makecell{}  & \makecell{} & \makecell{} & \makecell{0} \\

            \makecell{KAIST, 2019} & \makecell{\Checkmark} & \makecell{\Checkmark} & \makecell{} & \makecell{} & \makecell{} & \makecell{\Checkmark} & \makecell{} & \makecell{} & \makecell{\Checkmark} & \makecell{4}  & \makecell{\Checkmark} & \makecell{\Checkmark} & \makecell{0} \\

            \makecell{UrbanLoco, 2020} & \makecell{\Checkmark} & \makecell{} & \makecell{\Checkmark} & \makecell{} & \makecell{ } & \makecell{\Checkmark} & \makecell{} & \makecell{} & \makecell{\Checkmark} & \makecell{1}  & \makecell{} & \makecell{\Checkmark} & \makecell{3} \\
                
            \makecell{OpenLoris-Scene, 2020} & \makecell{\Checkmark} & \makecell{} & \makecell{} & \makecell{\Checkmark} & \makecell{ } & \makecell{\Checkmark} & \makecell{\Checkmark} & \makecell{} & \makecell{\Checkmark} & \makecell{}  & \makecell{\Checkmark} & \makecell{} & \makecell{9} \\
            
            \makecell{M2DGR, 2021} & \makecell{\Checkmark} & \makecell{\Checkmark} & \makecell{\Checkmark} & \makecell{} & \makecell{ } & \makecell{\Checkmark} & \makecell{} & \makecell{\Checkmark} & \makecell{\Checkmark} & \makecell{1}  & \makecell{} & \makecell{\Checkmark} & \rd{10} \\
            
            \makecell{FusionPortable, 2022} & \makecell{\Checkmark} & \makecell{} & \makecell{} & \makecell{} & \makecell{ } & \makecell{\Checkmark} & \makecell{} & \makecell{} & \makecell{\Checkmark} & \makecell{1}  & \makecell{} & \makecell{\Checkmark} & \makecell{5} \\         
            
            \makecell{Ground-Challenge, 2023} & \makecell{\Checkmark} & \makecell{\Checkmark} & \makecell{} & \makecell{\Checkmark} & \makecell{ } & \makecell{\Checkmark} & \makecell{\Checkmark} & \makecell{} & \makecell{\Checkmark} & \makecell{}  & \makecell{} & \makecell{} & \makecell{5} \\

            \makecell{M2DGR-Plus, 2024} & \makecell{\Checkmark} & \makecell{} & \makecell{\Checkmark} & \makecell{\Checkmark} & \makecell{ } & \makecell{\Checkmark} & \makecell{\Checkmark} & \makecell{} & \makecell{\Checkmark} & \makecell{1}  & \makecell{\Checkmark} & \makecell{\Checkmark} & \makecell{6} \\

            \makecell{MARS-LVIG, 2024} & \makecell{\Checkmark} & \makecell{\Checkmark} & \makecell{} & \makecell{} & \makecell{ } & \makecell{\Checkmark} & \makecell{} & \makecell{} & \makecell{\Checkmark} & \makecell{1}  & \makecell{} & \makecell{\Checkmark} & \makecell{6} \\

            \makecell{GrandTour, 2026} & \makecell{\Checkmark} & \makecell{\Checkmark} & \makecell{\Checkmark} & \makecell{} & \makecell{ } & \makecell{\Checkmark} & \makecell{\Checkmark} & \makecell{} & \makecell{\Checkmark} & \makecell{3}  & \makecell{} & \makecell{\Checkmark} & \nd\sl{52} \\
            
            \hline
            \makecell{\textbf{\mthreeDGR(Ours)}, 2026} & \makecell{\Checkmark} & \makecell{\Checkmark} & \makecell{\Checkmark} & \makecell{\Checkmark} & \makecell{ } & \makecell{\Checkmark} & \makecell{\Checkmark} & \makecell{\Checkmark} & \makecell{\Checkmark} & \makecell{2}  & \makecell{\Checkmark} & \makecell{\Checkmark} & {\fs \bf{68}} \\
            
            \hline
        \end{tabular}
    \end{adjustbox}
    \vskip 0.2em
    {\footnotesize\noindent\raggedright
    $^1$ is visual challenge, $^2$ is LiDAR degeneracy, $^3$ is GNSS denied zone, $^4$ is wheel slippage and $^5$ is omnidirectional camera.\par}
\vspace{-4mm}
    
\end{table}

\noindent\textbf{Sequence details.}
The real-world \mthreeDGR sequences cover routine operation and four transportation-relevant corner cases. For visual degradation, we consider dim lighting, time-varying illumination, dynamic foreground motion, and partial or complete occlusion. Indoors, darkness is emulated by switching off room lights and using a phone flashlight; illumination variation is created by periodically toggling the lighting condition; dynamic interference is induced by having a person move through the camera field of view; and occlusion is generated by intentionally blocking the lens. Outdoors, low-light and illumination-change sequences are recorded at night, while dynamic scenes include pedestrians, cyclists, and vehicles.

For LiDAR degeneracy, two sequence types emphasize poor geometric observability: a long corridor and a corridor-to-elevator transition. In both settings, the robot executes a loop and returns to the starting area, and ArUco-based start--end alignment is used to quantify accumulated drift caused by weak structural constraints. For wheel slippage, four corner cases are included: suspended-wheel ``float'' events, sharp turns, low-traction grass traversal, and rough roads with abrupt elevation changes. These sequences stress whether a fusion system can suppress corrupted wheel measurements before they bias the estimator. The GNSS-denial sequence starts with stable satellite reception, traverses a region without GNSS coverage, and returns to the initial area; ArUco markers make the induced drift observable even when absolute GNSS updates are absent.

Ground truth uses motion capture indoors at 360~Hz and RTK GNSS outdoors at 15~Hz, with ArUco-based relative alignment used for start--end drift evaluation in degeneracy and GNSS-denial cases. The \mthreeDGR Sim sequences provide simulator ground truth, including exact sensor extrinsics, and are grouped into Wild, Warehouse, and Tunnel scenes for controlled perturbation analysis. Table~\ref{tab:supp_m3dgr_sequence_overview} summarizes the number of sequences, distance, duration, storage size, and ground-truth source for each scenario category.

\begin{table}[!htbp]
        \scriptsize
        \caption{Overview of scenarios in the \mthreeDGR benchmark, including real-world and simulation sequences.}
        \centering
        \renewcommand{\arraystretch}{1.08}
        \setlength{\tabcolsep}{2pt}
        \label{tab:supp_m3dgr_sequence_overview}
        \begin{adjustbox}{width=0.9\linewidth}
        \begin{tabular}{*{19}c}
            \hline
            \multirow{2}{*}{\makecell{\textbf{Scenario}}} 
            & \multicolumn{5}{c}{\makecell{Visual Challenge}} 
            & \multicolumn{2}{c}{\makecell{LiDAR Degeneracy}}
            &
            & \multicolumn{4}{c}{\makecell{Wheel Slippage}}  
            & \multirow{2}{*}{\makecell{GNSS Denial}} 
            & \multirow{2}{*}{\makecell{Standard}}
            & \multicolumn{3}{c}{\makecell{Sim Data}}
            & \multirow{2}{*}{\makecell{TOTAL}} \\ 
            \cline{2-5} \cline{7-8} \cline{10-13} \cline{16-18}
            & Dark & VI$^1$ & Dynamic & Occlusion &  
            & Corridor & Elevator & 
            & WF$^2$ & ST$^3$ & Grass & RR$^4$ & &
            & Wild & Warehouse & Tunnel &\\   
            \hline
            \makecell{\textbf{Number}} & 5 & 4 & 3 & 4 &  & 2 & 1  & & 2 & 2 & 2 & 1 &  2 & 4 & 2 & 1 & 2 & 37 \\
            \makecell{\textbf{Dist/m}} & 1653.31 & 1055.58 & 355.97 & 1091.24 &  & 545.64 & 470.64 & & 101.55 & 170.88 & 318.91 & 457.35 & 1162.39 & 4485.49 & 159.991 & 10.123 & 861.257 & 12900.321 \\
            \makecell{\textbf{Duration/s}} & 2274 & 1458 & 609 & 1224 &  & 696 & 699 & & 171 & 238 & 459 & 533 & 1359 & 5101 & 254.980 & 194.990 & 1053.480 & 16324.45 \\
            \makecell{\textbf{Size/GB}} & 27.0 & 20.0 & 7.1 & 12.3 &  & 11.9 & 11.2 & & 3.3 & 2.9 & 9.7 & 10.4 & 23.2 & 86.0 & 14.4 & 2.6 & 60.4 & 302.4 \\
            \makecell{\textbf{Ground Truth}} & RTK/Mocap & RTK/Mocap & RTK/Mocap & RTK/Mocap & & ArUco & ArUco & & Mocap & Mocap & RTK & RTK & ArUco & RTK & Sim & Sim & Sim & ---- \\
            \hline
            \multicolumn{19}{l}{\footnotesize{$^1$ VI: varying illumination. $^2$ WF: wheel float. $^3$ ST: sharp turn. $^4$ RR: rough road.}}
        \end{tabular}
        \end{adjustbox}
        \vskip 0mm
\end{table}

\subsection{Evaluated Frameworks and Licenses}

Table~\ref{tab:supp_eval_frameworks} lists the frameworks evaluated in the M3DGR benchmark comparison of the main paper, together with their tested categories, public repository links, and associated open-source licenses when available. Methods follow the same order as Table~I in the main paper. 

\clearpage
\begin{table}[!htbp]
\tiny
\centering
\caption{Evaluated frameworks and their associated licenses.}
\label{tab:supp_eval_frameworks}
\renewcommand{\arraystretch}{0.94}
\setlength{\tabcolsep}{2pt}
\begin{adjustbox}{width=0.95\linewidth}
\begin{tabular}{p{0.24\linewidth} p{0.14\linewidth} p{0.40\linewidth} p{0.12\linewidth}}
\hline
\textbf{Method} & \textbf{Tested Category} & \textbf{Repository Link} & \textbf{License} \\
\hline

Ground-Fusion WIO & WIO & \url{https://github.com/SJTU-ViSYS/Ground-Fusion} & GPL-3.0 \\
GNSS SPP & GNSS reference & N/A & N/A \\
\ultrafusion~(WIO) & WIO & \url{https://github.com/sjtuyinjie/Ultra-Fusion} & MIT \\
ORB-SLAM2 & VO & \url{https://github.com/raulmur/ORB_SLAM2} & NOASSERTION \\
VINS-Mono & VIO & \url{https://github.com/HKUST-Aerial-Robotics/VINS-Mono} & GPL-3.0 \\
VINS-RGBD & RGBD-VIO & \url{https://github.com/Lab-of-AI-and-Robotics/VINS-RGBD} & GPL-3.0 \\
TartanVO & VO & \url{https://github.com/castacks/tartanvo} & BSD-3-Clause \\
ORB-SLAM3 & VO & \url{https://github.com/UZ-SLAMLab/ORB-SLAM3} & GPL-3.0 \\
VINS-GPS-Wheel & VIO+Wheel+GNSS & \url{https://github.com/Wallong/VINS-GPS-Wheel} & GPL-3.0 \\
DM-VIO & VIO & \url{https://github.com/lukasvst/dm-vio} & GPL-3.0 \\
GVINS & GVIO & \url{https://github.com/HKUST-Aerial-Robotics/GVINS} & GPL-3.0 \\
VIW-Fusion & VIW & \url{https://github.com/TouchDeeper/VIW-Fusion} & GPL-3.0 \\
Ground-Fusion & Multi-sensor & \url{https://github.com/SJTU-ViSYS/Ground-Fusion} & GPL-3.0 \\
MASt3R-SLAM & VO & \url{https://github.com/rmurai0610/MASt3R-SLAM} & NOASSERTION \\
\ultrafusion~(VIO) & VIO & \url{https://github.com/sjtuyinjie/Ultra-Fusion} & MIT \\
\ultrafusion~(VWIO) & VIW & \url{https://github.com/sjtuyinjie/Ultra-Fusion} & MIT \\
A-LOAM & LO & \url{https://github.com/HKUST-Aerial-Robotics/A-LOAM} & NOASSERTION \\
LeGO-LOAM & LO & \url{https://github.com/RobustFieldAutonomyLab/LeGO-LOAM} & BSD-3-Clause \\
LIO-mapping & LIO & \url{https://github.com/hyye/lio-mapping} & GPL-3.0 \\
LIO-SAM & LIO & \url{https://github.com/TixiaoShan/LIO-SAM} & BSD-3-Clause \\
LINS & LIO & \url{https://github.com/ChaoqinRobotics/LINS---LiDAR-inertial-SLAM} & NOASSERTION \\
LOAM-Livox & LO & \url{https://github.com/hku-mars/loam_livox} & GPL-2.0 \\
LiLi-OM & LIO & \url{https://github.com/KIT-ISAS/lili-om} & GPL-3.0 \\
LIO-Livox & LIO & \url{https://github.com/Livox-SDK/livox_mapping} & NOASSERTION \\
Faster-LIO & LIO & \url{https://github.com/gaoxiang12/faster-lio} & GPL-2.0 \\
IESKF-LIO & LIO & \url{https://github.com/chengwei0427/ESKF_LIO} & NOASSERTION \\
VoxelMap & LIO & \url{https://github.com/hku-mars/VoxelMap} & NOASSERTION \\
Fast-LIO2 & LIO & \url{https://github.com/hku-mars/FAST_LIO} & GPL-2.0 \\
CTLO & LO & \url{https://github.com/G3tupup/ctlo} & Unknown \\
Point-LIO & LIO & \url{https://github.com/hku-mars/Point-LIO} & Custom \\
LOG-LIO & LIO & \url{https://github.com/tiev-tongji/LOG-LIO} & GPL-2.0 \\
CT-LIO & LIO & \url{https://github.com/chengwei0427/ct-lio} & GPL-2.0 \\
DLIO & LIO & \url{https://github.com/vectr-ucla/direct_lidar_inertial_odometry} & MIT \\
HM-LIO & LIO & \url{https://github.com/chengwei0427/hm-lio} & NOASSERTION \\
KISS-ICP & LO & \url{https://github.com/PRBonn/kiss-icp} & MIT \\
SLICT & LIO & \url{https://github.com/brytsknguyen/slict} & GPL-2.0 \\
MM-LINS & LIO & \url{https://github.com/lian-yue0515/MM-LINS} & NOASSERTION \\
SLICT2 & LIO & \url{https://github.com/brytsknguyen/slict} & GPL-2.0 \\
PIN-SLAM & LiDAR-SLAM & \url{https://github.com/PRBonn/PIN_SLAM} & MIT \\
I2EKF-LO & LO & \url{https://github.com/YWL0720/I2EKF-LO} & GPL-2.0 \\
LTAOM & LO & \url{https://github.com/hku-mars/LTAOM} & NOASSERTION \\
LOG-LIO2 & LIO & \url{https://github.com/tiev-tongji/LOG-LIO2} & GPL-2.0 \\
Eq-LIO & LIO & \url{https://github.com/Eliaul/Eq-LIO} & NOASSERTION \\
Traj-LO & LO & \url{https://github.com/kevin2431/Traj-LO} & MIT \\
VoxelMap++ & LIO & \url{https://github.com/uestc-icsp/VoxelMapPlus_Public} & NOASSERTION \\
DMSA-SLAM & LiDAR-SLAM & \url{https://github.com/davidskdds/DMSA_LiDAR_SLAM} & MIT \\
Adaptive-LIO & LIO & \url{https://github.com/chengwei0427/Adaptive-LIO} & BSD-3-Clause \\
GLO & LO & \url{https://github.com/robosu12/GLO} & GPL-3.0 \\
LIGO & LIO+GNSS & \url{https://github.com/Joanna-HE/LIGO.} & BSD-3-Clause \\
CTE-MLO & (M)LO & \url{https://github.com/shenhm516/CTE-MLO} & GPL-2.0 \\
RKO-LIO & LIO & \url{https://github.com/PRBonn/rko_lio} & MIT \\
II-NVM & LIO & \url{https://github.com/chengwei0427/II-NVM} & NOASSERTION \\
Surfel-LIO & LIO & \url{https://github.com/93won/lidar_inertial_odometry} & MIT \\
GenZ-ICP & LO & \url{https://github.com/cocel-postech/genz-icp} & MIT \\
Voxel-SLAM & LiDAR-SLAM & \url{https://github.com/hku-mars/Voxel-SLAM} & GPL-2.0 \\
\ultrafusion~(LIO) & LIO & \url{https://github.com/sjtuyinjie/Ultra-Fusion} & MIT \\
\ultrafusion~(LWIO) & LIO+Wheel & \url{https://github.com/sjtuyinjie/Ultra-Fusion} & MIT \\
LVI-SAM & LIVO & \url{https://github.com/TixiaoShan/LVI-SAM} & BSD-3-Clause \\
R2LIVE & LIVO & \url{https://github.com/hku-mars/r2live} & GPL-2.0 \\
R3LIVE & LIVO & \url{https://github.com/hku-mars/r3live} & GPL-2.0 \\
Fast-LIVO & LIVO & \url{https://github.com/hku-mars/FAST-LIVO} & GPL-2.0 \\
Coco-LIC & LIVO & \url{https://github.com/APRIL-ZJU/Coco-LIC} & GPL-3.0 \\
SR-LIVO & LIVO & \url{https://github.com/ZikangYuan/sr_livo} & GPL-2.0 \\
Fast-LIVO2 & LIVO & \url{https://github.com/hku-mars/FAST-LIVO2} & GPL-2.0 \\
Ground-Fusion++ & Multi-sensor & \url{https://github.com/sjtuyinjie/Ground-Fusion2} & MIT \\
\ultrafusion~(LVIO) & LVIO & \url{https://github.com/sjtuyinjie/Ultra-Fusion} & MIT \\
\ultrafusion~(LVWIO) & LVWIO & \url{https://github.com/sjtuyinjie/Ultra-Fusion} & MIT \\

\hline
\end{tabular}
\end{adjustbox}
\end{table}

\vfill


 




\vfill

\end{document}